\documentclass[11pt]{article}

\usepackage[preprint]{acl}

\usepackage{times}
\usepackage{latexsym}

\usepackage[T2A,T1]{fontenc}

\usepackage[utf8]{inputenc}

\usepackage{microtype}

\usepackage{inconsolata}

\usepackage{graphicx}

\usepackage{tipa}
\usepackage{todonotes}
\usepackage{hyperref}
\usepackage{url}
\usepackage{wrapfig}
\usepackage{xspace}
\usepackage{graphicx}
\usepackage{algorithm}
\usepackage{algpseudocode}
\usepackage{arydshln}
\usepackage{amsmath,amsfonts,amssymb}
\usepackage{xcolor}
\usepackage{booktabs}
\usepackage{multirow}
\usepackage{tablefootnote}
\usepackage{refcount}
\usepackage{inconsolata}
\usepackage{pgfplots}
\usepackage{tikz}
\pgfplotsset{compat=1.18}
\usepackage{expex}

\DeclareRobustCommand{\legendbox}[1]{%
  \tikz[baseline=-0.6ex]\draw[draw=#1, fill=#1] (0,-0.15) rectangle (0.15,0.2);%
}
\definecolor{model1}{HTML}{1F77B4}
\definecolor{model2}{HTML}{FF7F0E}
\definecolor{model3}{HTML}{2CA02C}
\definecolor{model4}{HTML}{D62728}
\definecolor{model5}{HTML}{9467BD}
\definecolor{model6}{HTML}{8C564B}

\usepackage[utf8]{inputenc}
\usepackage{listings}

\definecolor{codegreen}{rgb}{0,0.6,0}
\definecolor{codegray}{rgb}{0.5,0.5,0.5}
\definecolor{codepurple}{rgb}{0.58,0,0.82}
\definecolor{backcolour}{rgb}{0.95,0.95,0.92}

\lstdefinestyle{mystyle}{
    backgroundcolor=\color{backcolour},
    commentstyle=\color{codegreen},
    keywordstyle=\color{magenta},
    numberstyle=\tiny\color{codegray},
    stringstyle=\color{codepurple},
    basicstyle=\ttfamily\footnotesize,
    breakatwhitespace=false,
    breaklines=true,
    captionpos=b,
    keepspaces=true,
    numbers=left,
    numbersep=5pt,
    showspaces=false,
    showstringspaces=false,
    showtabs=false,
    tabsize=2
}

\newcommand{\eng}{\textipa{N}}

\DeclareUnicodeCharacter{00F0}{\dh}
\DeclareUnicodeCharacter{014B}{\textipa{N}}
\DeclareUnicodeCharacter{0283}{\textesh}
\DeclareUnicodeCharacter{025B}{\textepsilon}
\DeclareUnicodeCharacter{0254}{\textopeno}
\DeclareUnicodeCharacter{028F}{\textscy}
\DeclareUnicodeCharacter{03B8}{\texttheta}
\DeclareUnicodeCharacter{026A}{\textsc{i}}
\DeclareUnicodeCharacter{028A}{\textupsilon}
\DeclareUnicodeCharacter{0259}{\textschwa}
\DeclareUnicodeCharacter{0268}{\textbari}
\DeclareUnicodeCharacter{028C}{\textturnv}
\DeclareUnicodeCharacter{0251}{\textscripta(A)}
\DeclareUnicodeCharacter{0292}{\textyogh}
\DeclareUnicodeCharacter{02B1}{\textsuperscript{\texthth}}
\DeclareUnicodeCharacter{03B2}{\textbeta}
\DeclareUnicodeCharacter{0263}{\textgamma}
\DeclareUnicodeCharacter{0261}{g}
\DeclareUnicodeCharacter{026C}{\textbeltl}
\DeclareUnicodeCharacter{0325}{\textsubring{r}}
\DeclareUnicodeCharacter{03C7}{\textchi} 

\usepackage{tcolorbox}
\usepackage{fancyvrb}
\usepackage{CJKutf8}
\usepackage[frozencache=true,cachedir=minted-cache]{minted}
\tcbuselibrary{breakable}
\tcbuselibrary{skins}

\newenvironment{promptbox}[2]{%
  \begin{tcolorbox}[
    breakable,
    enhanced,
    colback=gray!5,
    colframe=red!50!black,
    arc=0mm,
    title={#1},
    fonttitle=\bfseries,
    boxrule=0.5pt
  ]
  \begin{CJK}{UTF8}{min}
  \begin{flushleft}
  \small
  \VerbatimInput[fontsize=\small,breaklines=true]{#2}
  \end{flushleft}
  \end{CJK}
  \end{tcolorbox}
}{}

\usepackage{amsmath,amsfonts,bm}

\def\eqref#1{equation~\ref{#1}}

\def\1{\bm{1}}

\DeclareMathAlphabet{\mathsfit}{\encodingdefault}{\sfdefault}{m}{sl}
\SetMathAlphabet{\mathsfit}{bold}{\encodingdefault}{\sfdefault}{bx}{n}

\definecolor{lightgray}{rgb}{0.9, 0.9, 0.9}
\newcommand{\written}[1]{{\textlangle{#1}\textrangle}}
\newcommand{\dtoprule}{\specialrule{1pt}{0pt}{0.4pt}%
            \specialrule{0.3pt}{0pt}{\belowrulesep}%
            }
\newcommand{\dbottomrule}{\specialrule{0.3pt}{0pt}{0.4pt}%
            \specialrule{1pt}{0pt}{\belowrulesep}%
            }
\newcommand*\eiadfamily{\fontencoding{OT1}\fontfamily{eiad}\selectfont}
\newcommand*{\IASC}{{\eiadfamily{IASC}}}

\title{
Creating ConLangs to Probe the Metalinguistic Grammatical Knowledge of LLMs
}

\author{Chihiro Taguchi \\
  University of Notre Dame \\
  Notre Dame, IN, USA \\
  \texttt{ctaguchi@nd.edu} \\\And
  Richard Sproat \\
  Sakana AI \\
  Tokyo, Japan \\
  \texttt{rws@sakana.ai} \\}

\begin{document}
\maketitle

\begin{abstract}
We present a system that uses LLMs as a tool in the development of
Constructed Languages---ConLangs, which we call
{\IASC}\includegraphics[width=3mm]{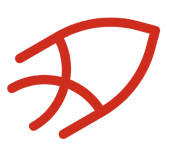} (Interactive Agentic
System for ConLangs). The system is modular in that it creates each
of the components---phonology, morphology and syntax, lexicon, orthography, and
grammatical handbook, using module-specific sets of prompts. The approach is
agentic in that various modules allow for refining the output given
automatically-generated commentary on a previous step. 
Our main goals are twofold.
First, we aim to provide tools that facilitate an engaging and enjoyable experience in creating artificially constructed languages.
Second, the focus of this paper is on using our ConLang framework as a novel way to explore what LLMs `know' about language---not what they
know about any particular language or encyclopedic facts, but how much they
know about and understand language and linguistic concepts. 
In the experiments, we particularly focus on the morphosyntax module and show that 
there is a fairly wide gulf in capabilities both among different LLMs and among
different linguistic specifications, with it being notably easier for systems to
deal with more typologically common patterns than rarer ones.
All code is released: \url{https://github.com/SakanaAI/IASC}.
\end{abstract}

\section{Introduction}
\label{sec:introduction}


The term `Constructed Language'---often shortened to `ConLang'---is used to refer to any artificially created language that is intended, in principle, to be as expressive as naturally evolved human languages.
The latter restriction is important since ConLangs are to be distinguished from artificial languages such as mathematical formulae and programming languages, which are artificial languages (at least in the formal sense of language) to be sure, but which are much more limited in the kinds of messages they can convey.

\begin{figure}
    \centering
    \includegraphics[width=1\linewidth]{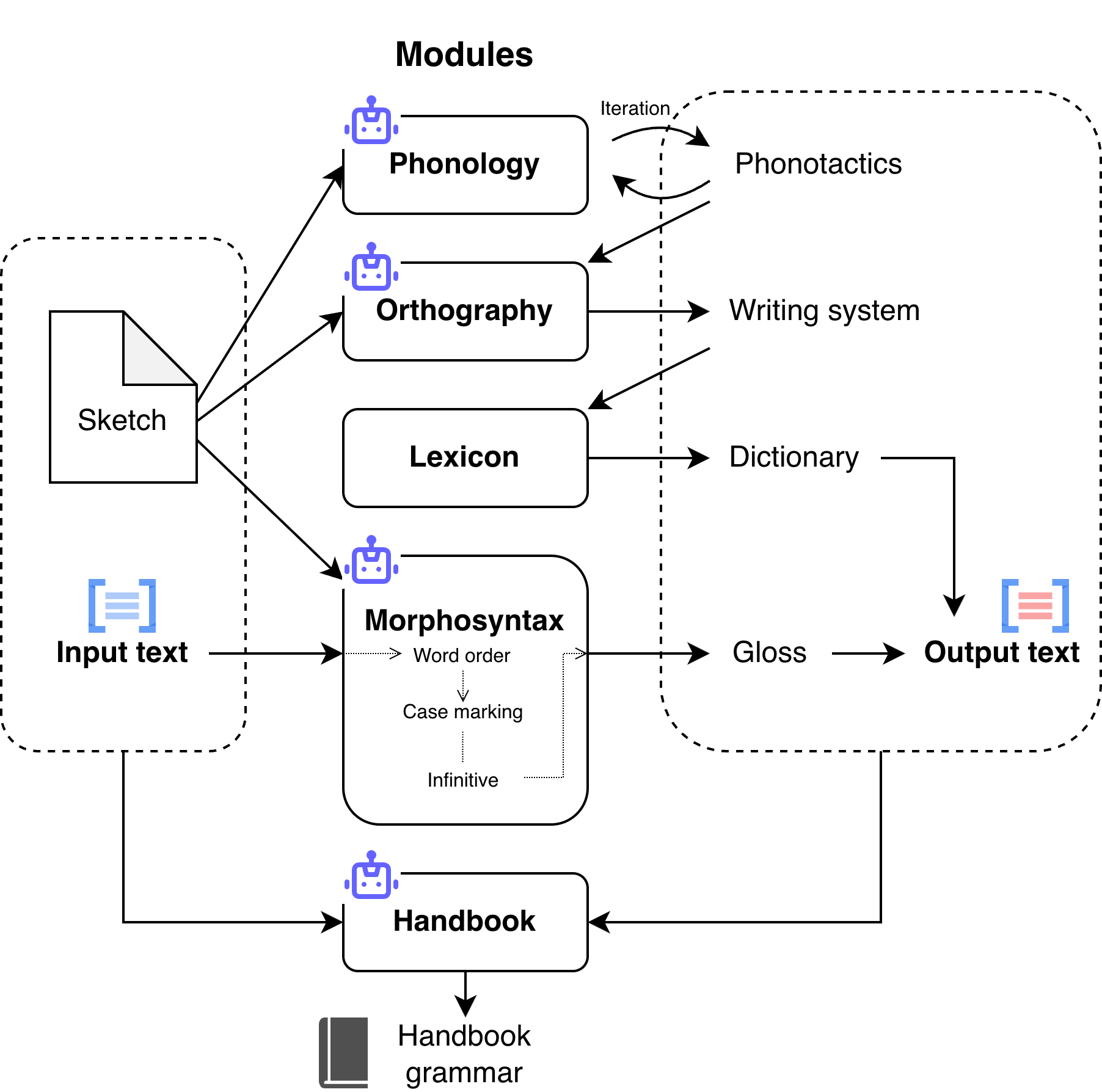}
    \caption{The workflow of {\IASC}\includegraphics[width=3mm]{Figures/redfish.png}.}
    \label{fig:conlang-workflow}
\end{figure}

This paper introduces {\IASC}\includegraphics[width=3mm]{Figures/redfish.png},\footnote{IASC is to be pronounced as /{\textprimstress}i{\textlengthmark}.ask/ in English. The word \textit{iasc} means `fish' in Modern Irish.} an \emph{Interactive Agentic System for ConLangs}, a system that enables users to customize linguistic parameters to design their own ConLang with a large language model (LLMs).
The system consists of modules for phonology, morphosyntax, lexicon creation, orthography, and grammatical handbook writing, with each module being responsible for generating linguistic rules or transformed output that follow high-level specifications given by the user.
For some modules, the output can be refined agentically using automatically generated feedback on a previous step.
The system's workflow is illustrated in Figure~\ref{fig:conlang-workflow}.

{\IASC} presents a unique challenge for LLMs as the workflow requires deep metalinguistic knowledge for each module.
That is, the model needs not only to know encyclopedic facts about languages and linguistic features but also to explicitly understand abstract linguistic concepts and manipulate language based on grammatical rules.
For example, suppose that a user wants to construct a language with a Verb-Subject-Object word order and has an input sentence ``the cat caught the fish.''
Then, the morphosyntax module of the system should ideally output ``caught the cat the fish.''
This sort of linguistic transformation requires strong metalinguistic reasoning skills beyond mere encyclopedic knowledge about languages.

Thus, the purpose of this paper is twofold.
First, we present {\IASC} as an engaging, flexible, and reproducible tool for designing and generating unique ConLangs.\footnote{In fact online tools to help with some of the steps already exist, including {\protect\url{http://VulgarLang.com}}, as well as instruction manuals such as \citet{Rosenfelder:12}. Our system can thus be thought of as a much broader, AI-driven version of these.}
Second, given the challenging nature of {\IASC}'s linguistic transformations, we use the morphosyntax module of the system as a benchmark to probe and evaluate the metalinguistic grammatical knowledge of LLMs.
Breaking down the problem into targeted tasks, such as setting word order and case marking, makes it easy to evaluate whether LLMs can be induced to produce outputs that conform to expectations.
For this task, we construct an evaluation dataset that tests nine different morphosyntactic configurations with typologically diverse sets of features.
As we will see, LLMs show a range of abilities that to some degree seem to relate to how frequent and well-documented a linguistic phenomenon is in the published literature, and how well the feature is likely to be supported in the system's training data.

\section{Related work}
\label{sec:related_work}

\subsection{ConLangs}
When people think of ConLangs, the first things that come to mind may be languages like Esperanto \cite{Zamenhof:1887} or Interlingua \cite{Gode:Blair:5191}, which are artificial languages designed to be like, and replace, natural languages, though with fewer of the grammatical complexities one often finds in real natural languages.
One might also think of fantasy languages, such as the languages of Tolkien's Middle-earth like Elvish (Quenya) \cite{Tolkien:1968}, or Science Fiction `alien' languages like Klingon \cite{Okrand:85} or the Heptapod language from \textit{Arrival} \cite{Villeneuve:16}.
Finally, there is a third category of languages that are supposed to be based on `logical' principles, such as LogLan or John Wilkins' `philosophical language' \citep{Wilkins:1668}.
In {\IASC}, we are interested in discovering to what degree Large Language Models (LLMs) can help in the creation of ConLangs that are much like real human languages in their properties, but are not intended to replace naturally evolved languages. Thus the ConLangs created here fit more into the fantasy-SF theme.

\subsection{LLMs' metalinguistic abilities}
A large amount of research has been devoted to investigating the linguistic
abilities of LLMs and their implications for linguistic theory. The latter point
in particular has included some strong claims, with \citet{Chomsky:EtAl:23}
arguing in an OpEd that LLMs are completely irrelevant to the study of language,
and \citet{Piantadosi:24} arguing the opposite, namely that LLMs completely
refute linguistic theories positing the innateness of grammatical knowledge.

Of most relevance to our work are studies on LLMs' metalinguistic knowledge, in
particular how good language models are at performing tasks that linguists might
perform, such as doing syntactic annotation, or correctly marking syntactic
arguments in a language that has a particular case-marking pattern.  Studies
along these lines include
\citep{Blevins:EtAl:23,Ginn:EtAl:24,Buhnila:25,Suvarna:EtAl:24,Jang:EtAl:25,%
  Merullo:EtAl:25,Dinu:EtAl:25,Saeed:EtAl:25,Pantelidou:EtAl:25,Begus:EtAl:25,%
  Goyal:Dan:25,Choudhary:EtAl:25}.

Studies of LLMs' abilities at constructing ConLangs have been more limited,
though artificial languages have figured in various studies of other aspects of
LLMs' linguistic abilities
\cite{White:Cotterell:21,Kallini:EtAl:24,Marmonier:EtAl:25,Swain:EtAl:25}.  Of
more direct relevance to our work, \citet{Diamond:23} investigated the use of
ChatGPT to construct languages via prompting, and studied properties of the
languages that the system came up with, and whether the constructed languages
obeyed the standard `Zipfian' power law distribution. He concludes that the
texts in the constructed languages show broadly similar word frequency distributions to their
English translations.

While we were conducting our research, \citet{Alper:EtAl:2025}'s paper on
ConlangCrafter appeared. Their model is similar in some respects to our approach
in that, unlike \citet{Diamond:23}, it breaks the problem down into separate
modules of phonology, morphology, syntax, lexicon generation and translation,
each of which is controlled by a high-level description of the properties the
target language should have. \citet{Alper:EtAl:2025}'s work is obviously related
to ours, but there are also important differences, particularly when it comes to
morphosyntactic feature generation. In particular, we adopt a very fine-grained
multistage approach to generating morphosyntax, which has the advantage that we
can investigate the question of how well LLMs `know' specific morphosyntactic
features, abstracting away from other features.

The interested reader can find a more detailed discussion of prior work in
Appendix~\ref{subsec:related_work_appendix}.

\section{Stages of ConLang Construction}
\label{sec:stages}

{\IASC}\includegraphics[width=3mm]{Figures/redfish.png} is a full system for
creating ConLangs, and as such consists of several components that each deal
with a different portion of the system. For reasons of space, we mainly focus in
this paper on the morphosyntactic component, which has the most implications for
what LLMs know about language. Other components, on which the interested reader
can find many more details in the Appendix, are phonotactics
(\ref{subsec:phonotactics}), lexicon generation (\ref{subsec:lexicon}),
orthography creation (\ref{subsec:orthography}), the writing of a grammatical
handbook of the language (\ref{subsec:handbook_generation}), and the translation
of further texts into the language (\ref{subsec:translation}). The system also
depends on a set of (English) texts that serve as the input from which target
language data is constructed: construction of this data is described in
\ref{subsec:stories}. In addition, in \ref{subapp:phonological_changes} we
discuss a novel agentic approach to generating phonological rules based on
attested sets of sound changes. And in \ref{subsec:low_resource} we discuss
possible applications of our approach to morphosyntax to low-resource languages.
Finally, \ref{subsec:linguistics} outlines the linguistic background that we
assume for our discussion.

However, to set the stage for the ensuing discussion, it is useful to present
briefly how the system constructs languages, and what an example language looks
like. Suppose we want a language with Welsh-like phonotactics, an
Ergative-Absolutive case system with a typologically unusual Object-Verb-Subject
word order like that of Mizo (see Section~\ref{sec:morphosyntax}), and an
orthography based on the Latin script. The system first creates a Python program
to generate morphemes that attempt to approximate the phonemes and their
phonotactic distribution for the target language. This is done iteratively, with
the system examining its own previous output, and possibly improving upon its
previous work in an agentic fashion. Next, the morphosyntax is developed, as
described in detail in the next section.  Once this is done, and based on the
morphosyntactic glosses of the source sentences for which we wish to create target-language
equivalents, the system will then create the needed target-language morphemes
using the phonotactic grammar to generate the forms. The system then develops an
orthography in the desired script, based on the phonotactic grammar developed
previously, and using a similar approach. Finally, the system writes a
grammatical handbook for the language, and invents a name for the language.

An  example sentence from the Welsh-Mizo-Latin ConLang `Thrlangai', is as follows:
%
%
\ex
\begingl
    \gla Chacho shitho krsüb nükhai næb rékólh there nonü noli snpor grnaim lauben. //
    \glb climate change country we ERG together address needs to with third singular //
    \glft `Our country needs to address climate change together' //
\endgl
\xe
Important features include ergative (ERG) case marking (\emph{næb}) on
\emph{krsüb nükhai} `our country', the placement of the object \emph{Chacho
shitho} `climate change' before the subject, conforming to the specifications of
the desired morphosyntactic structure, and the occurrence of the lateral
fricative /\textbeltl/, a very characteristic sound of Welsh, here written as
\written{lh} in the word \emph{rékólh}.

For another example, consider a language that has Japanese-like phonology,
Turkish-like morphosyntax---including Subject-Object-Verb word order and a
negation suffix on verbs---and a Cyrillic-based orthography. An example from this
language, called Koshanese, is as follows:
%
%
\ex
\begingl
    \glpreamble {\fontencoding{T2A}\selectfontКешизо дате шукутуа курауоотенсе} //
    \gla ke.ʃi-zo da.te ʃu.ku-tu-a ku.ra-u-o-o-ten-se //
    \glb she-NOM spicy food-SING-ACC like-NEG-PRES-IND-3SG-ACT //
    \glft `She doesn't like spicy food' //
\endgl
\xe
Examples of other generated languages can be found in the Appendix and at
\url{https://github.com/SakanaAI/IASC/tree/main/handbooks}.

\section{Morphosyntax}
\label{sec:morphosyntax}

This section details the Morphosyntax module of \IASC\ as well as experiments that probe LLMs' metalinguistic knowledge.
The term \emph{morphosyntax} denotes the subset of grammar that relates to how
words are arranged in a sentence (syntax) and how grammatical information is
marked on words (morphology).  Morphology addresses how a word is structured,
and syntax addresses how words form a phrase and a sentence.  For example, a
form \emph{dogs} consists of two \emph{morphemes} (minimum meaningful units):
\textit{dog} and the plural suffix \textit{-s}.  In this case, we can analyze the form
morphologically as \textit{dog-s}, and annotate it as dog-PLUR.  This
morpheme-to-morpheme annotation scheme, where morphemes like PLUR (plural marker) are
expressed in an abbreviated form, is called a \emph{gloss}.  Syntax, on the
other hand, is concerned with how words combine to form a larger phrase, such as
a noun phrase, prepositional phrase, or an entire sentence.  For example, in
English, a language with a relatively rigid word order, a canonical declarative sentence is
formed by arranging the constituents in the order of subject, verb, and object
(SVO), as in \emph{the dog chased the cat}.

What features are handled in morphology and syntax depend on languages.
Some languages employ rich morphology to encode various grammatical features.
For example, in Latin, nominal morphology encodes grammatical gender, number, and case, and verbal morphology encodes subject person, subject number, tense--aspect, mood, and voice.
On the other hand, other languages have simpler morphology and depend on syntactic constraints to encode grammatical features.
For example, though Vietnamese morphology does not encode any of the features
found in Latin morphology, at least some such information is nonetheless encoded in syntax by means of uninflected lexical items and rigid word order.
Given this cross-linguistic variability and the intertwined interactions in the realms of morphology and syntax, we treat them as one stage, \textit{i.e.} morphosyntax, in our ConLang construction pipeline.

\subsection{Research questions}
Of the modules in {\IASC}, we choose the Morphosyntax module as a probing system for evaluating the metalinguistic ability of LLMs since (1) this module requires a large amount of abstract understanding and reasoning about linguistic concepts, and (2) the fact that LLMs are trained primarily on text makes morphosyntactic knowledge a natural first thing to investigate over, say, phonology.

Morphosyntactic transformation of natural language text by a language model poses the following two research questions:
\begin{enumerate}
    \item \textbf{Metalinguistic ability}.
    It is already known that LLMs have linguistic knowledge \citep{Begus:EtAl:25} and are able to perform translation to/from an unseen low-resource language \citep{Tanzer:EtAl:24, zhu-etal-2025-evaluating}.
    It has also been reported that LLMs have the creative capacity to generate ConLangs on the fly \cite{Alper:EtAl:2025}.
    However, these findings still do not demonstrate the metalinguistic knowledge and ability of LLMs concerning morphosyntax.
    For instance, it is a straightforward task for linguistically-trained humans to transform an
    English sentence based on an instruction ``Change the English sentence to
    OSV word order'' or ``Change the English sentence so that the past tense is
    expressed by a prefix PAST-''. How well can an LLM perform such tasks?
    \item \textbf{Robustness to atypical typology}.
    How do atypical morphosyntactic features affect the ability of the LLM to perform the task? In principle, it is no harder to transform a sentence in SVO word order into, say OVS word order than it is to transform it to SOV word order. But the former is much less widely attested. Does this make a difference in the performance of LLMs on this task?
\end{enumerate}

\subsection{Methods}

In the Morphosyntax module, our goal is to manipulate the morphosyntactic structure of a source sentence with a language model  $\mathcal{M}$ such that the output follows the grammatical specifications of a given ConLang.
In particular, the input is a triple of a source sentence $s_\text{src}$, the
morphosyntactic parameters $G$ of our ConLang
grammar, and a prompt text $t$, and we wish to obtain the target sentence $s_\text{tgt} = \mathcal{M}(s_\text{src}, G, t)$.
In this Morphosyntax module, the output $s_\text{tgt}$ is not the final translation of the source sentence but is given in the form of a gloss.
For simplicity, the initial source sentence $s_\text{src}$ and the prompt $t$ are given in English.

Roughly speaking, there are two possible ways to implement the Morphosyntax module: (1) to transform a source sentence in one pass, and (2) to cumulatively transform a source sentence with iterative prompting.
Our preliminary experiments with the first approach showed that the language models struggle to follow
the grammar specifications, likely because the needed prompts became too long
and complicated.
For this reason, the implementations of our main experiments follow the cumulative morphosyntax approach.

In the Cumulative Morphosyntax approach, the source sentences are transformed through iterative prompting.
In each prompting step, the language model $\mathcal{M}$'s task is to transform the sentence of the previous output $s_{i-1}$ such that it follows the grammar specifications $G$ and the prompt text $t_i$ corresponding to a specific grammatical feature: $s_{i} = \mathcal{M}(s_{i-1}, G, t_i)$.
For example, suppose that we want to modify the case-marking system at the $i$th stage.
Then, the placeholders of the prompt text $t_i$ are filled based on the provided grammar $G$, and the model is expected to transform the input sentence $s_{i-1}$ in such a way that only the case-marking system of the sentence is modified in the output.

The features used in the experiments can be seen in
Table~\ref{tab:morphosyntax_features}.

\begin{table*}[t]
    \centering
    \tiny
    \setlength{\tabcolsep}{3pt}
    \begin{tabular}{@{}lllp{36em}@{}} \toprule
        Submodule & Features & Subfeatures & Values \\ \midrule
        Syntax & \texttt{main\_word\_order} & & \texttt{Literal["SOV", "SVO", "VSO", "VOS", "OSV", "OVS"]} \\
        & \texttt{oblique\_word\_order} & & \texttt{Literal["VOX", "VXO", "XOV", "XVO", "OVX", "OXV"]} \\
        & \texttt{adj\_noun\_word\_order} & & \texttt{Literal["NA", "AN]} \\
        & \texttt{posspron\_noun\_word\_order} & & \texttt{Literal["PossN", "NPoss"]} \\
        & \texttt{num\_noun\_word\_order} & & \texttt{Literal["NumN", "NNum"]} \\
        & \texttt{adposition\_noun\_word\_order} & & \texttt{Literal["PN", "NP"]} \\
        \midrule
        Morphology & \texttt{case} & \texttt{case\_marking} & \texttt{List[Literal["nominative", "accusative", "dative", "genitive", "ablative", "locative", "instrumental", "ergative", "absolutive"]]} \\
        & & \texttt{case\_marking\_strategy} & \texttt{Literal["prefix", "suffix", "prepositional word", "postpositional word"]} \\
        & & \texttt{oblique\_case\_marking} & \texttt{Literal["nominative", "accusative", "dative", "genitive", "ablative", "locative", "instrumental", "ergative", "absolutive"]} \\
        & \texttt{definiteness} & \texttt{definiteness} & \texttt{List[Literal["definite", "indefinite"]]} \\
        & & \texttt{definiteness\_marking\_strategy} & \texttt{Literal["prefix", "suffix", "prepositional word", "postpositional word"]} \\
        & \texttt{adjective\_agreement} & \texttt{adjective\_agreement} & \texttt{List[Literal["number", "case", "definiteness"]]} \\
        & & \texttt{adjective\_agreement\_strategy} & \texttt{Literal["prefix", "suffix", "prepositional word", "postpositional word"]} \\
        & \texttt{comparative} & \texttt{comparative} & \texttt{List[Literal["comparative", "superlative", "equative"]]} \\
        & & \texttt{comparative\_marking\_strategy} & \texttt{Literal["prefix", "suffix", "prepositional word", "postpositional word"]} \\
        & \texttt{tense\_aspect} & \texttt{tense\_aspect} & \texttt{List[Literal["present", "past", "future", "perfect", "imperfect", "immediate past", "recent past", "remote past", "nonpast"]]} \\
        & \texttt{nominal\_number} & \texttt{nominal\_number} & \texttt{Literal[List["singular", "plural", "dual", "paucal"]]} \\
        & & \texttt{nominal\_number\_marking\_strategy} & \texttt{Literal["prefix", "suffix", "prepositional word", "postpositional word"]} \\
        & & \texttt{tense\_aspect\_marking\_strategy} & \texttt{Literal["prefix", "suffix", "prepositional word", "postpositional word"]} \\
        & \texttt{person} & \texttt{person\_agreement} & \texttt{List[Literal["first", "second", "third"]]} \\
        & & \texttt{person\_marking\_strategy} & \texttt{Literal["prefix", "suffix", "prepositional word", "postpositional word"]} \\
        & & \texttt{verbal\_number\_agreement} & \texttt{List[Literal["singular", "plural", "dual", "paucal"]]} \\
        & & \texttt{verbal\_number\_marking\_strategy} & \texttt{Literal["prefix", "suffix", "prepositional word", "postpositional word"]} \\
        & \texttt{voice} & \texttt{voice} & \texttt{List[Literal["active", "passive"]]} \\
        & & \texttt{voice\_marking\_strategy} & \texttt{Literal["prefix", "suffix", "prepositional word", "postpositional word"]} \\
        & \texttt{mood} & \texttt{mood} & \texttt{List[Literal["indicative", "subjunctive", "imperative", "conditional"]]} \\
        & & \texttt{mood\_marking\_strategy} & \texttt{Literal["prefix", "suffix", "prepositional word", "postpositional word"]} \\
        & \texttt{relativization} & \texttt{relativization\_order} & \texttt{Literal["head-initial", "head-final"]} \\
        & & \texttt{relativization\_marking} & \texttt{Literal["head-marking", "dependent-marking"]} \\
        & & \texttt{relativizer\_position} & \texttt{Literal["prepositional", "postpositional"]} \\
        & & \texttt{relativizer\_morpheme} & \texttt{Literal["affix", "word"]} \\
        & \texttt{infinitive} & \texttt{infinitive} & \texttt{Literal["infinitive"]} \\
        & & \texttt{infinitive\_marking\_strategy} & \texttt{Literal["prefix", "suffix", "prepositional word", "postpositional word"]} \\
        & \texttt{nominal\_number} & \texttt{nominal\_number} & \texttt{List[Literal["singular", "plural", "dual", "paucal"]]} \\
        & & \texttt{nominal\_number\_marking\_strategy} & \texttt{Literal["prefix", "suffix", "prepositional word", "postpositional word"]} \\
        & \texttt{inclusive\_exclusive} & & \texttt{bool} \\
        \bottomrule
    \end{tabular}
    \caption{The details of the Morphosyntax data structure.
    \texttt{Literal[...]} means that one value must be chosen from the list, and \texttt{List[Literal[...]]} means that multiple values from the list are accepted.}
    \label{tab:morphosyntax_features}
\end{table*}

\subsection{Evaluation data}
To quantify the performance of the LLM's morphosyntactic manipulation, we constructed an evaluation dataset.\footnote{The dataset is publicly available under the CC BY-SA 4.0 license at \url{https://huggingface.co/datasets/ctaguchi/conlang_eval_dataset}.}
The dataset consists of 45 source sentences
and nine ConLang feature sets, amounting to 405 sentences.
These source sentences are designed to test a variety of grammatical features.
For the ConLang feature sets, typologically diverse combinations of features were selected.
To reflect the real-world typological diversity, we prepare sets of features inspired by eight existing languages: Arabic, Fijian, French, Hixkaryana, Mizo, Turkish, Vietnamese, and Welsh.
In this paper, these feature sets are referred to by their quoted lowercase counterparts (\textit{e.g.}, `arabic'), but note that these names are only used for convenience and \emph{do not strictly reflect all morphosyntax details of the named language}.
We also prepare a feature set with a typologically extremely rare combination of the features, which we call the `hard' language.
This language contains an atypical mixture of head-initialness and
head-finalness as well as complex and unusual morphology.
See Appendix~\ref{sec:morphosyntax-params} for the full details of the morphosyntax feature parameters.
The gold data was created by one of the authors---a trained linguist----by translating source sentences to the targets in the form of glosses.

\subsection{Evaluation metrics}
The evaluation stage of the Morphosyntax module receives the gloss transformed from the source sentence as the input and measures the correctness.
Since there can be small non-deterministic variations in the output formats provided by the language models, the output translations are fed into another lightweight language model to obtain a unified structured output in JSON format.
In our experiments, GPT 4.1 mini was used for this structuring step.

\paragraph{Translation Edit Rate (TER).}
TER was proposed as a metric of translation quality by \cite{Snover:EtAl:06} and measures the correctness of a translation output by simply dividing the number of edits by the reference length:

$$
\text{TER} = \dfrac{\text{Number of edits}}{\text{Reference length}}.
$$

In contrast to edit distance-based metrics such as WER, in TER a shift of a group of words by any number of positions counts as 1 edit.
For example, compare the reference and hypothesis sentences given below:

\begin{itemize}
    \setlength\itemsep{0em}
    \item Reference: \texttt{the cat caught the fat rat}
    \item Hypothesis: \texttt{the fat rat caught the cat}
\end{itemize}

In this case, TER counts two edit operations to get the reference from the hypothesis: \texttt{the fat rat} to the right and \texttt{the cat} to the left.
Therefore, the TER score for the hypothesis is $2/6 \simeq 0.167$.

\paragraph{Stem Edit Rate (SER).}
SER is an extension of TER to measure the correctness of the positions of the \emph{stems} in the hypothesis glosses.
This metric is useful to measure how well the language model is able to manipulate syntax, regardless of morphological changes in the output.
In our implementation, SER is calculated by (1) obtaining the stems through
removing the morphological features, which are expressed using capital letters and
(2) calculating the TER of the stems.
That is, TER counts the edits of the exact forms including the morphological feature tags, and SER only counts the edits of the stems, ignoring the morphological features.

\paragraph{Morphological Feature Error Rate (MFER).}
In contrast to SER, MFER measures how well the language model can gloss morphological features in the correct form in the correct position.
Thus, this metric does not take word order into account;
rather, for each reference word, it searches for a word in the hypothesis that minimally deviates from the reference word.
The morpheme-level edit distance is then used to compute the error rate.
Note that, while we take into account whether a morpheme is a prefix or a suffix, we do not take into account the order within the sequence of prefixes or suffixes.
In other words, sequences of suffixes like \texttt{-PRES-3SGNOM} and \texttt{-3SGNOM-PRES} are considered identical when computing MFER.
It is implemented this way because the order within a sequence of affixes is not explicitly specified in the ConLang prompts.

\paragraph{Morphosyntactic Error Rate (MSER).}
MSER is proposed to take both morphological errors and syntactic errors into account by calculating a weighted average of SER and MFER ($\alpha = 0.5$ by default):
$$
\text{MSER} = \alpha \cdot \text{SER} + (1 - \alpha) \text{MFER}.
$$

\paragraph{Lemma F1 score (Lem F1).}
This metric measures how well the model is capable of lemmatizing the input sentences. Lemmatization is important, since if one is to transform the source sentences into target glosses, one needs the base form of the original words, not the inflected form: thus \texttt{dog-PL-ERG} rather than \texttt{dogs-PL-ERG}. 
In preliminary experiments we found that several language models failed to lemmatize correctly, thereby negatively affecting other metric scores such as SER.  

Given a predicted sequence $s_p$ and its reference sequence $s_r$, we obtain the frequencies of the lemmata $l_p$ and $l_r$ expressed as multisets (bags-of-words).
We then count the true positives $\text{TP} = | l_p \cap l_r|$ (\textit{i.e.}, the predicted lemmata also included in the reference lemmata), false positives $\text{FP} = l_p \setminus l_r$, and false negatives $\text{FN} = l_r \setminus l_p$.
After this, the lemma precision ($\text{LemPrec}$) and the lemma recall ($\text{LemRec}$) are calculated as:
$$
\text{LemPrec} = \dfrac{\text{TP}}{\text{TP} + \text{FP}},~~~
\text{LemRec} = \dfrac{\text{TP}}{\text{TP} + \text{FN}}.
$$
The lemma F1 score ($\text{LemF}_1$) is the harmonic mean of the lemma precision and the lemma recall; namely,
$$
\text{LemF}_1 = \dfrac{2 \cdot \text{LemPrec} \cdot \text{LemRec}}{\text{LemPrec} + \text{LemRec}}.
$$
In the evaluation below, we utilize the macro lemma F1 scores (\textit{i.e.}, averaging over the F1 scores of each prediction).

\subsection{Evaluation results}
We evaluated the morphosyntactic transformation performance across six different language models of varying quality: Claude 3.5 Sonnet, Gemini 2.5 Flash, Gemini 2.5 Pro, GPT-4o-mini, GPT-5-mini, and GPT-5.
The temperature is set to 0 where it can be specified, and thinking is enabled for the reasoning models.
For snapshots of the specific versions, see Table~\ref{tab:llms}.
Figure~\ref{fig:mser-results} summarizes a comparison of MSER scores across the 9 feature sets and 6 LLMs.
For full results see Table~\ref{tab:full-results} in Appendix~\ref{sec:morphosyntax-full-results}.
Each result is obtained from a single run.

\begin{table}[t]
    \centering
    \small
    \setlength{\tabcolsep}{4pt}
    \begin{tabular}{@{}ll@{}} \toprule
        Model & Version \\ \midrule
        gpt-4o-mini & gpt-4o-mini-2024-07-18 \\
        gpt-4.1-mini & gpt-4.1-mini-2025-04-14 \\
        gpt-5-mini & gpt-5-mini-2025-08-07 \\
        gpt-5 & gpt-5-2025-08-07 \\
        gemini-2.5-flash & gemini-2.5-flash (released: June 17, 2025) \\
        gemini-2.5-pro & gemini-2.5-pro (released: June 17, 2025) \\
        claude-3-5-sonnet & claude-3-5-sonnet-20240620 \\
        \bottomrule
    \end{tabular}
    \caption{Models and their snapshots.}
    \label{tab:llms}
\end{table}




\begin{figure*}
\centering
\begin{tikzpicture}
\begin{axis}[
    width=16.40cm,
    height=5.90cm,
    ybar,
    bar width=5.50pt,
    ymin=0.0,
    ymax=100,
    unbounded coords=jump,
    enlarge x limits=0.08,
    ylabel={MSER},
    ylabel shift=-8pt,
    xlabel={Language},
    xlabel shift=-4pt,
    title={},
    symbolic x coords={{ arabic,fijian,french,hard,hixkaryana,mizo,turkish,vietnamese,welsh }},
    xtick=data,
    x tick label style={rotate=30, anchor=east, font=\fontsize{9}{11}\selectfont},
    y tick label style={font=\fontsize{9}{11}\selectfont},
    ylabel style={font=\fontsize{10}{12}\selectfont},
    xlabel style={font=\fontsize{10}{10}\selectfont},
]

\addplot+[ybar, draw=model1, fill=model1, bar shift=-13.75pt] coordinates {
    (arabic,13.282981)
    (fijian,14.363771)
    (french,17.297397)
    (hard,46.800303)
    (hixkaryana,45.952392)
    (mizo,54.293002)
    (turkish,13.673490)
    (vietnamese,16.960765)
    (welsh,4.485364)
};
\addplot+[ybar, draw=model2, fill=model2, bar shift=-8.25pt] coordinates {
    (arabic,20.956596)
    (fijian,8.836689)
    (french,17.571733)
    (hard,42.415047)
    (hixkaryana,17.009320)
    (mizo,8.592279)
    (turkish,41.154296)
    (vietnamese,21.958333)
    (welsh,11.210571)
};
\addplot+[ybar, draw=model3, fill=model3, bar shift=-2.75pt] coordinates {
    (arabic,6.902357)
    (fijian,20.113342)
    (french,4.294171)
    (hard,30.867212)
    (hixkaryana,30.706488)
    (mizo,4.665738)
    (turkish,26.097803)
    (vietnamese,6.369427)
    (welsh,2.071429)
};
\addplot+[ybar, draw=model4, fill=model4, bar shift=2.75pt] coordinates {
    (arabic,78.276846)
    (fijian,51.997035)
    (french,52.338457)
    (hard,81.875325)
    (hixkaryana,78.812745)
    (mizo,77.218964)
    (turkish,68.948106)
    (vietnamese,57.433844)
    (welsh,66.395073)
};
\addplot+[ybar, draw=model5, fill=model5, bar shift=8.25pt] coordinates {
    (arabic,12.731771)
    (fijian,104.057565)
    (french,64.401165)
    (hard,52.221014)
    (hixkaryana,46.320459)
    (mizo,78.129027)
    (turkish,67.981244)
    (vietnamese,54.864776)
    (welsh,8.879781)
};
\addplot+[ybar, draw=model6, fill=model6, bar shift=13.75pt] coordinates {
    (arabic,20.959513)
    (fijian,26.826620)
    (french,5.189394)
    (hard,59.188235)
    (hixkaryana,32.643315)
    (mizo,20.446402)
    (turkish,67.928592)
    (vietnamese,37.821062)
    (welsh,42.193418)
};

\end{axis}
\end{tikzpicture}
\caption{MSER by model and language.
All scores are the results with the few-shot in-context examples.
\legendbox{model1} Claude 3.5 Sonnet,\quad
\legendbox{model2} Gemini 2.5 Flash,\quad
\legendbox{model3} Gemini 2.5 Pro,\quad
\legendbox{model4} GPT-4o-mini,\quad
\legendbox{model5} GPT-5-mini,\quad
\legendbox{model6} GPT-5.}
\label{fig:mser-results}
\end{figure*}



\paragraph{Typological unusualness.}
As the results in Table~\ref{tab:full-results} demonstrate, all of the language models struggled to transform the sentences into the `hard' language.
The `hard' language has a rich morphology, and typologically unusual features
such as case prefixation.
In addition, we can observe higher SERs in languages with a typologically unusual
word order such as `fijian' (VOS), `mizo' (OSV), and `hard' (OSV).
In contrast, languages with a typologically more common word order, such as `french' (SVO), `arabic' (VSO), and `welsh' (VSO), yielded lower errors in general.
Notably, Claude 3.5 Sonnet, Gemini 2.5 Pro, and GPT-5-mini yielded perfect SER scores ($0.00$) in arabic and welsh, both of which have a VSO word order.
When we classify SOV, SVO, and VSO as common word orders and the others as uncommon word orders, there is a weak significant correlation between word order and SER (point-biserial correlation coefficient $r_{pb} = 0.29$, $p = 0.0026 < 0.05$).

\paragraph{Lemmatization: analytic vs. synthetic language.}
Another interesting observation is that some high-performance LLMs struggled to correctly lemmatize analytic languages, while performing better in synthetic languages with heavy morphological inflection.
Let us suppose that `fijian', `mizo', and `vietnamese' are analytic, and the other languages are synthetic, based on the description given in \cite{wals-22}.
Then, we observe a weak significant correlation between the inflectional degrees and the Lemma F1 ($r_{pb} = 0.32$, $p = 0.00082 < 0.05$).
In other words, languages that rely more on word order and function words rather than inflectional morphology generally posed greater challenges for lemma recognition by LLMs.
In the output for the analytic languages, we observed that some inflected nouns and verbs were left inflected.
Overall, the language that the models struggled the most is `hard' with an average MSER score of 85.21, and the easiest language was `welsh' with an average MSER score of 28.26.

\paragraph{Effect of few-shot in-context learning.}

Given the difficulty of the task, we have also tested the models' performance with an additional stage called \emph{review}, where a model is provided with a prompt that contains a summary of the grammar and four example sentences with label glosses at the end of the Morphosyntax module.
This allows the model to digest the grammar with a few-shot in-context learning
(ICL) setting.
In this case, as shown in Table~\ref{tab:morph-results-icl}, Claude 3.5 Sonnet,
Gemini 2.5 Flash, Gemini 2.5 Pro, and GPT-5 showed an improvement for most metrics.

\begin{table}[t]
\centering
\setlength{\tabcolsep}{3pt}
\small
\begin{tabular}{@{}llrrrrr@{}} \toprule
    Model & ICL& TER& SER& MFER& MSER& LemF1 \\ \midrule
    {\scriptsize claude-3-5-sonnet} & No & 77.78 & 33.37 & 41.73 & 37.55 & 76.17 \\
        & Yes & \textbf{55.22} & \textbf{26.29} & \textbf{24.18} & \textbf{25.23} & \textbf{89.13} \\
    \midrule
    {\scriptsize gemini-2.5-flash} & No & 74.86 & 43.07 & 47.64 & 45.36 & 73.11 \\
        & Yes & \textbf{48.35} & \textbf{18.02} & \textbf{24.14} & \textbf{21.08} & \textbf{89.99} \\
    \midrule
    {\scriptsize gemini-2.5-pro} & No & 83.00 & 60.80 & 47.57 & 54.18 & 74.38 \\
        & Yes & \textbf{40.94} & \textbf{12.64} & \textbf{16.71} & \textbf{14.68} & \textbf{89.30} \\
    \midrule
    {\scriptsize gpt-4o-mini} & No & \textbf{87.58} & 71.64 & 66.67 & 69.16 & 52.79 \\
        & Yes & 88.83 & \textbf{69.93} & \textbf{66.36} & \textbf{68.14} & \textbf{53.49} \\
    \midrule
    {\scriptsize gpt-5-mini} & No & \textbf{80.80} & \textbf{62.70} & 47.80 & 55.25 & 68.28 \\
        & Yes & 84.76 & 62.79 & \textbf{46.01} & \textbf{54.40} & \textbf{69.13} \\
    \midrule
    {\scriptsize gpt-5} & No & 78.53 & 45.58 & 46.26 & 45.92 & 61.76 \\
        & Yes & \textbf{57.57} & \textbf{35.70} & \textbf{33.90} & \textbf{34.80} & \textbf{79.90} \\
    \bottomrule
\end{tabular}
\caption{Average scores over languages by models comparing performance with and without a few-shot in-context learning (ICL) stage.}
\label{tab:morph-results-icl}
\end{table}

\begin{table*}[ht]
    \centering
    \small
    \begin{tabular}{@{}lll@{}} \toprule
        Model & ICL & Output \\ \midrule
        Source & & The children played in the park. \\
        Gold & & DEF-ABS-PLUR-child DEF-LOC-SING-park RECPAST-3PLABS-IND-play. \\
        \midrule
        claude-3-5-sonnet & No & DEF-PLUR-ABS-child DEF DEF-SING-park-LOC DEF in RECPAST-ACT-play-3PLNOM. \\
        & Yes & ABS-DEF-PLUR-child LOC-DEF-SING-park the in RECPAST-3PLABS-IND-ACT-play. \\ \cmidrule{2-3}
        gemini-2.5-flash & No & abs-DEF-PLUR-child active-plural-play-third LOC-DEF-SING-park in. \\
        & Yes & LOC-DEF-SING-park in ERG-DEF-PLUR-child RECPAST-3PLERG-IND-ACT-play. \\ \cmidrule{2-3}
        gemini-2.5-pro & No & abs-DEF-plur-child loc-DEF-sing-park in plural-RECPAST-active-play-third. \\
        & Yes & DEF-ABS-PLUR-child DEF-LOC-SING-park in RECPAST-3PLABS-IND-ACT-play. \\ \cmidrule{2-3}
        gpt-4o-mini & No & park the in played children the. \\
        & Yes &  park the in played children the. \\ \cmidrule{2-3}
        gpt-5-mini & No & ABS-the DEF-PLUR-child PL-ACT-PL-REMPAST-IND-play INS-DEF-SING-park the in. \\
        & Yes & INS-DEF-SING-park in ABS-DEF-PLUR-child ACT-PLUR-REMPAST-IND-play-THIRD. \\ \cmidrule{2-3}
        gpt-5 & No & (NA) \\
        & Yes & RECPAST-3PLABS-IND-ACT-play DEF-ABS-PLUR-child DEF-INS-SING-park in. \\
        \bottomrule
    \end{tabular}
    \caption{A comparison of example model outputs in the `hard' language.
    The most plausible output is that of Gemini 2.5 Pro, which correctly follows the word order, ergative-absolutive alignment, and most of the morphology.
    GPT-5 without in-context examples did not generate any output for the sentence.}
    \label{tab:morph-results-example}
\end{table*}



\paragraph{Model `intelligence' and performance.}
The results demonstrate that there are differences in the ability of morphosyntactic manipulation across different feature sets and language models.
It can be observed that the LLMs are generally able to transform the input into ConLang grammars to some extent.
However, there is a significant difference between larger LLMs
(Claude 3.5 Sonnet, Gemini 2.5 Pro, and GPT-5) and the others ($r_{pb} = -0.31$, $p = 0.001 < 0.05$).
As the results so far suggest, not all models were able to perform the task satisfying the conditions given in the prompt instructions.
In particular, GPT-4o-mini and GPT-5-mini struggled with the task even when the few-shot examples were given.
The model that performed the best in this task overall was Gemini-2.5-Pro with few-shot ICL, outperforming the other models in all the metrics but Lemma F1 by a significant margin as demonstrated in Table~\ref{tab:morph-results-icl}.

Table~\ref{tab:morph-results-example} presents a comparison of the actual outputs by the different models for a sentence in the `hard' language with few-shot ICL.
Overall, Gemini 2.5 Pro is able to follow the grammar specifications of the `hard' language, including the OSV word order, ergative-absolutive alignment, and prefixation.
On the other hand, GPT-4o-mini completely ignores morphology even with the few-shot examples.
These results suggest that larger state-of-the-art LLMs do have relevant
metalinguistic knowledge, though capabilities still greatly vary across models.

\section{Conclusion}
This study introduced a novel agentic LLM system for ConLang generation.
With this system, we examined how effectively LLMs can manipulate and generate morphosyntactic features
in response to explicit instructions.
We found that larger SOTA LLMs are mostly able to follow the grammatical
instructions correctly to generate appropriate target-language glossing, while
smaller models struggled to perform the task.  The performance of the larger
LLMs can be further improved by providing few-shot in-context examples before
generating the final output.  We observed correlations between
the LLMs' performance and typological features,  LLMs struggling
more with ConLangs that have a unusual word order or have no
morphological inflection.

To specifically answer our research questions:

\begin{enumerate}
  \item \textbf{Metalinguistic ability}.
  Larger LLMs are generally capable of word order permutations, though there is a definite performance drop with typologically unusual orders.
  \item \textbf{Robustness to atypical typology}.
  Larger LLMs performed well at many of the tasks, though even some large LLMs struggled with the `hard' language, which combined several typologically unusual configurations.
  Oddly, LLMs performed worse on a crucial processing step---lemmatization---if the target language was more `analytic'.
\end{enumerate}

For the large proprietary models in particular, we cannot unfortunately know
\emph{why} the models performed as they did. However, it is unlikely that they
would have been exposed to much text in languages that have, say, VOS
orders. Similarly, while the models have presumably seen grammatical information
(\textit{e.g.} on Wikipedia) about such languages, such grammars will be rare compared to
grammars for languages with more `typical' grammatical properties.  But in order
to perform the transformation to that word order, it is not necessary that the
model have seen real life instances: it suffices that it understands concepts
like \emph{subject}, \emph{object}, and \emph{verb}, and that it understands what it means to
put phrases in a particular order. For this sort of metalinguistic task, the
`knowledge' of the LLMs seems to be mostly sufficient, though
once again, we observe that the more unusual the target, the more the model
struggles with the task.

\clearpage
\clearpage
\section{Limitations}
\label{sec:limitations}

There are many obvious limitations to our work.

The first and most obvious one is that in developing the morphosyntactic
properties of the target ConLang, we assume a model where one translates in
successive steps into the target language.  The result is that the target
language will likely retain properties of the source language---English, in our
experiments. In the case of English this means that one will often find residues
of English-specific syntax, such as \emph{do}-support, use of
auxiliary \emph{be} in passives, along with direct translations of English
expressions, in the target language.  One possible approach to improve this
would be to explicitly instruct the model to avoid direct translation of
particular terms (especially things like auxiliary verbs) and be creative in
its choice of target expressions.

The morphosyntax is currently limited to the set of features that we have
selected as interesting, and there are many features that we have not treated,
that are grammatically expressed in at least some languages. One example is
diminutives and augmentatives, found in many languages.  Some of the features
that we do in principle cover, such as gender, remain problematic.  If one is creating a
ConLang, except in the case of natural gender, it is not obvious how to assign
gender to a given target word.  Beyond that, the current system only supports
masculine, feminine and neuter genders, which reflects just one kind of gender
system.  Obviously this could be extended to cover, for example, Bantu-like
gender systems, but the question of how the system would decide what gender to
assign to a given word remains. 

Our results are of course also limited by our evaluation set which, while 
we attempted to make it comprehensive, obviously could be expanded to cover a wider range of examples 
that test the systems' abilities.

Another issue is that the morphology is currently limited to prefixation and
suffixation. Furthermore, once an affix has been chosen to express a particular
morphosyntactic feature, the same affix will be used to express that feature in
all cases. Sequences of morphosyntactic features such as SING-NOM are expressed
by sequences of affixes. The current system is thus biased towards a strictly
affixal, single-exponent, agglutinative strategy, where furthermore there is no
paradigmatic variation.

The phonotactic, phonological and orthographic components of our system, along
with the generation of handbooks as documented in the Appendix also have
limitations, but we omit discussion of these since they are not part of the main
paper.

Finally, we compared several LLMs in their ability to carry out the tasks
required of them. On balance one system that fared very well over a variety of
tasks was Gemini 2.5 Pro. But we hasten to add that our coverage is not
comprehensive. In any case, the abilities of LLMs are constantly changing and
improving, so it is likely that the capability of systems to carry out the
detailed instructions required of them in our tasks will only improve over time.

\section{Ethical considerations}
\label{sec:ethics}

While it is hard to think of any technology that could not be used for harm, and
while it is certainly the case that LLMs and Generative AI more generally have raised
serious red flags when it comes to ethical and safety issues, we believe that
our work does not present any obvious ethical problems.  All of the data used in
this project are either public data or synthetic, and all of the data created
are, by definition, synthetic.

Since the process of ConLang creation sketched here is in part random, it is
possible that the system might generate, unintentionally, outputs that could be
deemed offensive. The most obvious example would be words that are homophones or
close homophones with offensive terms in some language. In that situation, a
user of our system could easily intervene and change the offensive item, if so
desired.

During the preparation of this paper, AI assistants were used to correct grammatical errors and improve readability, as one of the authors is not a native speaker of English.
\section{Acknowledgments}
\label{sec:acknowledgments}

We thank various colleagues at Sakana.ai for feedback, in particular Robert
Lange for some in-depth discussions about agentic systems. We also thank three
anonymous reviewers and the metareviewer for helpful feedback.

\clearpage

\bibliography{refs}

\appendix
\clearpage
\newpage

\section{Appendix}
\label{app:appendix}

\subsection{A more detailed discussion of related work}
\label{subsec:related_work_appendix}

\subsubsection{Prior work on probing LLMs' linguistic abilities}
\label{subsubsec:prior_linguistic}


Since the advent of transformer-based large language models, there has been a
large amount of work probing the linguistic abilities of these models. A lot of
this work investigates the abilities of the models on linguistic constructions
in particular languages, or probing LLMs to discover how and what linguistic
information is represented; some recent work along these lines includes
\cite{Nikolaev:Pado:23,Katinskaia:Yangarber:24,Qiu:EtAl:24,Ide:EtAl:25,Tjuatja:EtAl:25}. Other
work makes bolder claims about LLMs' linguistic abilities, with
\citet{Piantadosi:24} in particular arguing that these models refute Chomsky's
theory of the innateness of grammatical knowledge---and see
\citet{Kodner:EtAl:23} for a response. Given that one clear difference between
humans' and LLMs' learning of language is the amount of data involved, of
particular relevance is work in the BabyLM challenge task
\citep{Warstadt:EtAl:23} (and see also \citet{Warstadt:Bowman:24}), where
language models have been shown to produce competitive performance with much
more human-sized amounts of data. \citet{Collado-Montanez:EtAl:25} argue
linguistic competence of LLMs does not scale as steeply with model size as does
factual knowledge, and that linguistic competence can be decoupled from LLMs'
more general knowledge, and handled with a smaller model, decoupled from the
model of general knowledge. Also related is work such as that of
\citet{Beuls:VanEecke:24} that investigates more plausible language-learning
scenarios that involve agents communicating with each other in a simulated
social environment more akin to the way humans learn
language. \citet{Kallini:EtAl:24}, add further fuel to the fire by arguing that
structurally `impossible' languages are much less easily learned by LLMs than
are structurally human-like languages, countering a claim by
\citet{Chomsky:EtAl:23} that such models do not distinguish between observed and
unobserved linguistic structure, and suggesting that whatever learning biases
humans may have, LLMs seem to have them too.  \citet{Hu:EtAl:24} argue that
LLMs' judgments align well with those of humans on some important grammatical
constructions (and see \citet{Cho:25} for recent work on LLMs' pragmatic
competence). \citet{Lopez-Otal:EtAl:25} present a literature review of 160
research works that seek to probe how linguistic knowledge of various kinds is
represented in transformer-based models.  There has also been interest in
investigating to what degree the representations arrived at by LLMs mirror what
one can infer from neurology, for example via fMRI: see \citet{Parra:25} for a
recent example.  Finally, see \citet{Lappin:24} for a balanced assessment of
arguments that have been made about LLMs' abilities, and lack thereof, and
\citet{Futrell:Mahowald:25} for an equally balanced assessment of the
implications of LLMs for the field of linguistics.

Of more direct relevance to our work are studies that investigate metalinguistic
knowledge, in particular how good language models are at performing tasks that
linguists might perform, such as performing syntactic annotation, or correctly
marking syntactic arguments in a language that has a particular case-marking
pattern. Thus \citet{Blevins:EtAl:23} investigate the abilities of language
models to perform various sorts of structural analysis. \citet{Ginn:EtAl:24}
investigate LLMs' ability to gloss low-resource languages, showing that without
any further training or fine-tuning, such models perform better than
transformers trained specifically for these tasks. \citet{Buhnila:25}
investigates the ability of smaller ({\textasciitilde}3B-parameter) models for
tasks related to pragmatic implicature.
On the phonological side,
\citet{Suvarna:EtAl:24} present the PhonologyBench benchmark that tests LLMs'
abilities at such tasks as rhyme generation or syllable counting, and
demonstrate that LLMs show striking performance on many tasks, despite receiving
no explicit exposure to phonology, yet at the same time fall short of human
abilities in this area. See also \citet{Jang:EtAl:25}, who refine prompting to
improve LLMs' performance on the PhonologyBench benchmark. A related study is \citet{Merullo:EtAl:25}, who
investigate the rhyming abilities of Llama-3.2-1B-Instruct, suggesting that the
LLM forms an internal representation of phonology that, however, differs in some
notable ways from those that humans seem to construct.\footnote{Needless to say,
phonological knowledge can benefit from having access to speech. See
\citet{deHeerKloots:Zuidema:24} for a study of human-like biases in neural
speech models.}  \citet{Cho:25a} shows that LLMs seem to use phonological cues
similar to those used by humans in determining the gender in Korean given names.
\citet{Dinu:EtAl:25} investigate LLMs' ability to engage in
tasks that involve linguistic creativity, such as creating new words based on
existing derivational and compounding mechanisms, and argue that LLMs actually
outperform humans in many of these tasks.  \citet{Saeed:EtAl:25} investigate
LLMs' abilities at inflection morphology across a wide range of languages
showing, not surprisingly, that while such models perform well on English, they
do not deal so well with uncommon morphological patterns. They further show
Chain-of-Thought and other `thinking' strategies of recent models can actually
degrade performance on these tasks. However, whether it is the morphological patterns per se,
or merely the frequency with which they occur in the data is unclear: \citet{Pantelidou:EtAl:25},
in a controlled study involving predicting inflected forms of novel words in a variety of
languages, argue that the main driver of differences in performance is the amount
of data available for the language.
Finally, \citet{Begus:EtAl:25} investigate
LLMs' abilities at various metalinguistic tasks such as determining if a
sentence is ambiguous, or identifying syntactic recursion. They particularly
single out OpenAI's o1 model for vastly outperforming other models, and point to
its Chain-of-Thought reasoning abilities as one possible reason, suggesting that
such reasoning approaches may be of help in at least some metalinguistic tasks.

\citet{Goyal:Dan:25} and \citet{Choudhary:EtAl:25} study the performance of
LLMs on Linguistics Olympiad puzzles. \citeauthor{Goyal:Dan:25} show that LLMs
struggle to handle many types of Olympiad problems, in particular ones that
require the participant to extract rules. \citeauthor{Choudhary:EtAl:25} observe,
again unsurprisingly, that the models show a strong English bias in that they do
better on Olympiad questions that probe features that happen to coincide with
properties of English. Complex morphology also posed problems for models, with
the models performing worse on problems that involved larger numbers of
morphological features. Models were helped in such problems by making the
morpheme boundaries in examples explicit. Some of these conclusions relate to
results we report in this paper. However one important difference between what we
present and prior work is that our approach to generating ConLangs involves
generating abstract English-glossed morphosyntactic annotations, rather than
surface forms directly. This allows us to probe the question of how well LLMs
understand the various \emph{concepts} underlying different morphosyntactic
feature settings, abstracting away from particulars of how those settings are
marked.

\subsubsection{Prior work on ConLangs}
\label{subsubsec:prior_conlangs}

Prior work on testing the linguistic abilities of LLMs has made use of
artificially constructed languages. For example \citet{White:Cotterell:21}
construct artificial languages specifically designed to allow one to investigate
the sensitivity of the LLM to the variation of one feature, such as word order.
\citet{Kallini:EtAl:24} construct `impossible' languages---languages that have
features that would be exceedingly unusual in any natural language---to test
whether these unusual features affect LLMs' abilities to learn the
languages. And \citet{Marmonier:EtAl:25} use artificially perturbed versions of
real languages, along with grammatical descriptions of those languages to test
whether LLMs can benefit from descriptions contained in grammar books, inspired
by work of \citet{Tanzer:EtAl:24} and others. Very recent work by
\citet{Swain:EtAl:25} uses a simple artificial constructed formal language
`Tinkatongue' to investigate LLMs' abilities to adapt to a new language.

But to date, there has been very little research on testing LLMs' abilities to
\emph{construct} ConLangs---this despite a long-term appeal of ConLangs to the
general public \citep{Okrent:09,Piperski:17}. One exception to this is work of
\citet{Diamond:23}---who uses the term `GenLangs' in his paper. The author used
ChatGPT, and prompted the model to name a new language, then create text in that
language, along with an English translation. The model was then asked to write a
short story plus its translation. Finally, the model was asked to write as long
a story as it could in the language. Diamond examined the grammatical properties
of the three created languages, noting that they differed in, for example, how
verbs conjugate: in one language verbs were conjugated for tense and aspect as
well as mood, whereas in another language verbs did not conjugate at all. One of
the languages chose a base-twenty number system, something that occurs in about
21\% of the world's languages \citep{Comrie:WALS:13}, but would have been
significantly \emph{underrepresented} in the training data that the LLM was
exposed to. One language chose nominal case prefixes, which is typologically
very unusual. Diamond's main goal though was to examine whether the corpora
generated for the GenLangs obeyed the standard `Zipf's Law' power law
distribution. He argues that English text generated by the LLM has a
distribution very close to that of natural text, especially when compared to
LLM-generated gibberish.\footnote{Note though that LLM-generated texts are
well-known to display platykurtosis, exhibiting relatively few outliers compared
to natural text \citep{Badurowicz:EtAl:24}, so that with a large enough amount of text
one would be able to detect a difference between natural English, and
LLM-generated English.}  The GenLangs showed more variation, with some languages
having distributions close to their English translations, but with one with a
decidedly different distribution (though still different from gibberish).

In his conclusions, Diamond notes that `practical' ConLangs of the first type we
outlined above (such as Esperanto) took many years of human effort to
develop. He contrasts this with the possibility of generating languages using
AI: ``For the first time in history, we have the opportunity to discover what
happens when artificial languages are developed by artificial intelligence.''
While Diamond's work is a proof of concept that creation of an AI-generated
ConLang in one fell swoop is a possibility, for reasons we have outlined,
in this paper we take a different approach.

As we were preparing this paper, a short paper appeared by
\citet{Alper:EtAl:2025} describing their ConlangCrafter system. Their model is
similar in some respects to our approach in that, unlike
\cite{Diamond:23}, it breaks the problem down into separate modules of
phonology, morphology, syntax, lexicon generation and translation. Their system
starts with a high-level description of the language to be generated, which can
be as specific or inspecific as the user wants. The system then creates each
component according to the specifications, refining each stage with internal
feedback to make each component more consistent: for example if the language to
be generated should be VSO, an SOV sentence would make the system less
consistent than if all sentences are consistently VSO. The output is a system
that is then used to translate a set of 10 test sentences from English into the
target language. They evaluate the systems for Diversity, Language Consistency
and Translation Consistency.  \emph{Diversity} is calculated by having the
system construct a set of languages, encoding each language as a one-hot vector
over WALS features \citep{Comrie:WALS:13}, and computing the pairwise Hamming
distance. Note that one of \citet{Alper:EtAl:2025}'s goals is to create a system
that allows for as diverse a set of ConLangs as possible.  \emph{Language
Consistency}, ranging between 0 and 1, computes the proportion of generated
sentences deemed consistent, and \emph{Translation Consistency} similarly
measures consistency for the translations. For the largest model they
used---DeepSeek-R1 they report improvements in each of these measures compared
to a baseline using the same model, where the baseline is ``a naive single-stage
generation method'' that ``attempts to generate a full language sketch and
translations of given sentences in our format with a single prompt, without
multi-hop reasoning or iterative self-refinement'' (page 4). For their smaller
models, Gemini-2.5-Flash and Gemini-2.5-Pro, the baseline seemed to outperform
the more elaborate model at least for the Language Consistency measure---see
their Table~1, page 5. While \citet{Alper:EtAl:2025}'s work is obviously related
to ours, there are also important differences, particularly when it comes to
morphosyntactic feature generation. As we outline in this paper, we adopt a very
fine-grained multistage approach to generating morphosyntax, which has the
advantage that we can investigate the question of how well LLMs `know'
specific morphosyntactic features, abstracting away from other features.

\subsection{A Quick Intro to (Relevant Bits of) Linguistics}
\label{subsec:linguistics}

A lot of the discussion in this paper assumes some familiarity with linguistics,
at least basic terminology and concepts. In this section we introduce concepts
that we will assume that the reader will have or will have gained some
familiarity with. Further details of most of the terminology that we will assume
can be found in standard textbooks such as \cite{Akmajian:EtAl:01}, and of
course there are many good Wikipedia pages that cover the topics in
depth. The World Atlas of Language Structures \citep{WALS} gives detailed
examples of many ways in which languages vary.

We will assume familiarity with concepts like \emph{phoneme}, \emph{morpheme}
and \emph{affix}. In many of the ConLangs, nouns, verbs and adjectives may
inflect for things like \emph{case}, \emph{number}, \emph{person}, \emph{tense}
and \emph{aspect}.  Within each of the above categories, languages make
different choices as to what kind of information to mark.

Within case, a fundamental distinction is between languages that are
\emph{nominative}-\emph{accusative} versus those that are
\emph{ergative}-\emph{absolutive}. In the former, any subject of the sentence
would be marked in the nominative case, and direct objects would be marked
accusative. In the latter, subjects of \emph{transitive} sentences are marked
ergative, whereas subjects of \emph{intransitive} sentences, along with objects,
are marked absolutive.

For number, many languages mark a distinction between \emph{singular} and
\emph{plural}, but some languages also mark other numbers, such as \emph{dual}
for instances involving two referents.

\emph{First}, \emph{second} and \emph{third person} distinctions are marked in a
majority of languages, but languages also make finer grained distinctions. One
notable one is in the \emph{first person plural}, which in some languages is
divided between \emph{inclusive}---including the hearer---and \emph{exclusive},
not including the hearer.

Languages mark various kinds of tense information on verbs, a common pattern
being \emph{past}, \emph{present} and \emph{future}. But languages may also mark
various degrees of remoteness in the past tense (distant past, very distant
past, etc.).

These are some of the kinds of variation we are interested in modeling. In
any case we introduce examples of phenomena we are
modeling where they arise.

\subsection{Phonotactics}
\label{subsec:phonotactics}

\subsubsection{Modeling phonotactics}
\label{subsubsec:phonotactics}

The first stage in the construction of a language is to decide what the basic
morphemes of the language should look like.  Do they involve relatively simple
syllables, as in for example Hawaiian where each syllable consists of a single
consonant and one or two vowels (cf. well-known Hawaiian names like \emph{Mauna
Kea})? Or does it allow much more complex syllables, as for example Russian,
where complex syllables like \emph{vzgl'ad} `view', are common. And what are the
basic set of sounds of the language? Again, Hawaiian has just twelve basic
sounds, whereas English, for example has on the order of 50.  Linguists
typically refer to the latter issue as the \emph{phoneme inventory} of the
language, and the first as the \emph{phonotactics}.

In this discussion we will conflate both of these under the rubric of
\emph{phonotactics}, and present an agentic prompting-based method for
determining an initial set of morpheme shapes, and for refining that initial
proposal.

We start with a language upon which to model the phonotactics of our
ConLang. The initial prompt schema can be found in
Appendix~\ref{subsubapp:phonotactics_prompt}. The prompt requests the LLM to
create a standalone Python program that will generate morphemes that are similar
in terms of phoneme inventory and phonotactics to a given target language. The
code should include two global dictionaries named \verb|consonants| and
\verb|vowels|, which indicate the relative frequencies of the different
sounds. There should be a function \verb|generate_morpheme| which will randomly
generate one morpheme that conforms to the phonotactics.

With the target language of Welsh, for example, the following output is created
(LLM=Claude~Sonnet~3.5):

\begin{lstlisting}[language=Python,inputencoding=utf8]
import random
import argparse

# Welsh-like phoneme set with relative frequencies
consonants = {
    "p": 3, "t": 5, "k": 4, "b": 2, "d": 3, "g": 2,
    "m": 3, "n": 7, "ŋ": 1, "f": 2, "v": 1, "θ": 1,
    "ð": 1, "s": 4, "ʃ": 2, "h": 2, "χ": 1, "l": 4,
    "r": 3, "j": 2, "w": 2
}

vowels = {
    "a": 8, "ɛ": 6, "e": 5, "i": 6, "ɪ": 4, "ɔ": 5,
    "o": 5, "u": 4, "ʊ": 3, "ə": 7, "ɨ": 2
}

# Morpheme templates (C = consonant, V = vowel)
templates = [
    "CV", "CVC", "CVCC", "CCVC", "CCVCC",
    "CV.CV", "CVC.CV", "CV.CVC", "CVC.CVC", "CCVC.CV", "CV.CCVC"
]

def choose_phoneme(phoneme_dict):
    return random.choices(list(phoneme_dict.keys()), weights=list(phoneme_dict.values()))[0]

def generate_morpheme():
    template = random.choice(templates)
    morpheme = []
    for syllable in template.split('.'):
        syll = []
        for char in syllable:
            if char == 'C':
                syll.append(choose_phoneme(consonants))
            elif char == 'V':
                syll.append(choose_phoneme(vowels))
        morpheme.append(' '.join(syll))
    return ' . '.join(morpheme)

def main():
    parser = argparse.ArgumentParser(description='Generate Welsh-like ConLang morphemes')
    parser.add_argument('--num_morphemes', type=int, default=10, help='Number of morphemes to generate')
    args = parser.parse_args()

    for _ in range(args.num_morphemes):
        print(generate_morpheme())

if __name__ == "__main__":
    main()
\end{lstlisting}

The process is then iterated. The LLM is prompted with the original prompt, the
code it produced, and some sample output from the code that it wrote in the
first phase (or the error message produced if the code failed). It is also
instructed to pay attention to the following:

\begin{lstlisting}[language=bash]
There may also be phonemes that are particularly characteristic of the language
that are missing from your previous code, or ones that should not be
there. Please also check for that.
\end{lstlisting}

The full listing for a prompt generated for the second phase, with the target
language Welsh, can be found in
Appendix~\ref{subsubapp:phonotactics_improvement_prompt}.

This prompt causes the system to improve a number of aspects, most notably
adding phonemes that were missing in the first round.  A phoneme that is
particularly characteristic of Welsh is the voiceless lateral fricative
/\textbeltl/. This was missing from the initial set, but was added in the
second:

\begin{lstlisting}[language=Python,inputencoding=utf8]
import random
import argparse

# Updated Welsh-like phoneme set with relative frequencies
consonants = {
    "p": 3, "t": 5, "k": 4, "b": 2, "d": 3, "g": 2,
    "m": 3, "n": 7, "ŋ": 1, "f": 2, "v": 1, "θ": 1,
    "ð": 1, "s": 4, "ʃ": 2, "h": 2, "χ": 1, "l": 4,
    "r": 3, "j": 2, "w": 2, "ɬ": 2  # Added 'ɬ' (voiceless alveolar lateral fricative)
}

vowels = {
    "a": 8, "ɛ": 6, "e": 5, "i": 6, "ɪ": 4, "ɔ": 5,
    "o": 5, "u": 4, "ʊ": 3, "ə": 7, "ɨ": 2, "æ": 3  # Added 'æ'
}

# Updated morpheme templates (C = consonant, V = vowel)
templates = [
    "CV", "CVC", "CVCC", "CCV", "CCVC",
    "CV.CV", "CVC.CV", "CV.CVC", "CVC.CVC", "CCV.CV", "CV.CCV"
]

# Valid initial consonant clusters
initial_clusters = [
    "pl", "pr", "bl", "br", "tr", "dr", "kl", "kr", "gl", "gr",
    "fl", "fr", "θr", "sp", "st", "sk", "sm", "sn", "sf", "ʃr"
]

# Valid final consonant clusters
final_clusters = [
    "lp", "lb", "lt", "ld", "lk", "lg", "lm", "ln", "lf", "lv", "lθ", "ls",
    "rp", "rb", "rt", "rd", "rk", "rg", "rm", "rn", "rf", "rv", "rθ", "rs",
    "mp", "nt", "ŋk", "st", "ft", "θt", "sp", "sk"
]

def choose_phoneme(phoneme_dict):
    return random.choices(list(phoneme_dict.keys()), weights=list(phoneme_dict.values()))[0]

def generate_consonant_cluster(clusters):
    return random.choice(clusters)

def generate_morpheme():
    template = random.choice(templates)
    morpheme = []
    for syllable in template.split('.'):
        syll = []
        if syllable.startswith('CC'):
            syll.append(generate_consonant_cluster(initial_clusters))
            syllable = syllable[2:]
        for char in syllable:
            if char == 'C':
                if syll and syll[-1] in consonants:
                    syll.append(choose_phoneme(vowels))
                else:
                    syll.append(choose_phoneme(consonants))
            elif char == 'V':
                syll.append(choose_phoneme(vowels))
        if syllable.endswith('CC'):
            syll[-2:] = [generate_consonant_cluster(final_clusters)]
        morpheme.append(' '.join(syll))
    return ' . '.join(morpheme)

def main():
    parser = argparse.ArgumentParser(description='Generate Welsh-like ConLang morphemes')
    parser.add_argument('--num_morphemes', type=int, default=10, help='Number of morphemes to generate')
    args = parser.parse_args()

    for _ in range(args.num_morphemes):
        print(generate_morpheme())

if __name__ == "__main__":
    main()
\end{lstlisting}

In the second iteration, the system also added initial and final consonant
clusters, which were absent from the initial system, providing a
\verb|generate_consonant_cluster| function for this purpose Unfortunately in so
doing, it violated the requirement laid out in the initial instructions:
``Within each morpheme, please place spaces between every phoneme.''  This error
was pointed out in the prompt in the third iteration:

\begin{lstlisting}[language=bash]

1. The following combined phonemes suggest that your code is not correctly
dealing with spaces between phones. Please correct this:

lf, lm, fl, sm, θr, gl, ʃr, rb, lθ, lt, lg, sf, rp, ft, rn, fr, pl, bl, lk, rf,
pr, sn, lv, rs, lp, rv, rθ, rd, gr, rm, rg, ln, sp, rk, sk, kr, ls
\end{lstlisting}

The system corrected this in the third iteration by correcting this function

\begin{lstlisting}[language=bash]
1. In the `generate_consonant_cluster` function, I've added space separation
   between the consonants in the cluster:
   ```python
   return ' '.join(list(random.choice(clusters)))
   ```
This change ensures that all phonemes, including those in consonant clusters,
are properly space-separated.
\end{lstlisting}

Further iterations refine various other aspects of the model.

Since the goal is to create a language that has a phonotactics that is
\emph{similar} to that of the target language, we do not expect that the system
will produce an exact match. Nonetheless, if we specify Welsh as the language
upon which to model the system, one would hope that the system would produce
something that is at least closer to Welsh than, say, Russian. One way to
measure this is to compute the perplexity of a sample of the words generated by
the system against simple ngram language models trained on Wiktionary data for a
variety of languages. Unfortunately this is not a perfect solution since there
is a wide range of levels of phonetic transcription in Wiktionary, with data for
some languages (e.g. Japanese), showing exceedingly fine-grained phonetic
transcriptions, whereas for other languages, such as French, the transcription
is much more coarse-grained. Thus a `Japanese'-based phonotactics that produces
reasonably plausible output such as the following:
\begin{quotation}
\noindent t\textesh\ i  .  k i\\
d o\\
k a a\\
e\\
\textfishhookr\ a u\\
a\\
m i  .  b u\\
\textfishhookr\ u e  .  k i\\
\textesh\ e i  .  s e\\
s i  .  s a  .  k o\\
s o  .  d a  .  n i\\
\end{quotation}
will nonetheless fail to match well against a model trained on the Wiktionary
entries for Japanese. We discuss this directly below in Section~\ref{subsubsec:phonotactics_results}.

Of course, phonology involves more than phonotactics. In ongoing work, we
develop a novel approach to creating plausible phonological rule sets. While
this is not yet integrated into the system, we describe the approach in
Appendix~\ref{subapp:phonological_changes}, and release the code along with the
rest of the software.

\subsubsection{Evaluating Phonotactics}
\label{subsubsec:phonotactics_results}

If one asks for a language that is like a given target language in terms of its
phonotactics, how much like that language is it? One way to measure this is to
compare it against a traditional n-gram language model constructed over IPA
transcriptions for the target language from Wiktionary. To this end, we used the
OpenGrm Ngram library \citep{Roark:EtAl:12} to build 3-gram language models for
each of 221 languages from Wiktionary. Of these, 49 languages have at least
10,000 entries, and we used this subset in the experiment below. We also
filtered by the proportion of OOV symbols, omitting languages for which the OOV
rate was 20\% or higher.

We then use the final phonotactic model constructed using the procedure
described above, and used it to generate 10,000
morphemes from the language, then computed its perplexity given the 3-gram
language model.  Table~\ref{tab:phonotactics_results} shows each of the target
languages, the top pick in terms of (lowest) perplexity given the 3-gram
language model, and whether or not the correct language occurs in the top ten
(lowest ten in terms of perplexity). Four LLMs were used: Claude 3.5 Sonnet,
GPT-4o 2024-08-06, Qwen2.5 72B Instruct and Llama 3.3 70B Instruct.  As can be
seen from the table, for Claude, in 4 out of 10 cases the target language has
the lowest perplexity, and in 5 out of ten cases the correct language is within
the top 10. The other languages did not fare so well. However, all other LLMs
performed worse than Claude.
\begin{table*}[ht]
\begin{center}
{\footnotesize
\begin{tabular}{rrrrrrrrr}
\toprule
 & \multicolumn{2}{c}{Claude} & \multicolumn{2}{c}{GPT} & \multicolumn{2}{c}{Qwen}  & \multicolumn{2}{c}{Llama} \\
Target & Top & Top~10 & Top & Top~10 & Top & Top~10 & Top & Top~10\\
\midrule
English & \textbf{English} & $\checkmark$ & \textbf{English} & $\checkmark$ & \textbf{English} & $\checkmark$
& Anc. Greek & $\checkmark$ \\
French & \textbf{French} & $\checkmark$ & Turkish & $\checkmark$  & Turkish & $\checkmark$
& Armenian & $\checkmark$ \\
German & \textbf{German} & $\checkmark$ & Anc. Greek &  & English & $\checkmark$
& Maltese & \\
Hawaiian & Basque &  & Romanian & & Romanian &
& Armenian & \\
Japanese & Turkish &  & Romanian & & Romanian &
& Armenian & \\
Old~English & Turkish &  & Turkish & & Anc. Greek &
& Turkish & \\
Russian & French &  & Turkish & & Turkish &
& Indonesian & \\
Spanish & Turkish & $\checkmark$  & Romanian & & Turkish &
& Arabic & \\
Turkish & \textbf{Turkish} & $\checkmark$ & \textbf{Turkish} & $\checkmark$  & Esperanto & $\checkmark$
& Romanian & $\checkmark$  \\
Welsh & English &  & Turkish & & Catalan &
& Tagalog & \\
\bottomrule
\end{tabular}
} 
\caption{\label{tab:phonotactics_results}Results comparing lowest perplexity
  match for ten target language phonotactics against 3-gram language models
  built on Wiktionary IPA transcriptions for languages with at least 10K
  Wiktionary entries. Shown are the predicted (lowest-perplexity) language, and
  whether the correct language is within the top 10. LLMs used were Claude 3.5
  Sonnet, GPT-4o 2024-08-06, Qwen2.5 72B Instruct and Llama 3.3 70B Instruct.}
\end{center}
\end{table*}

In some cases the differences can be explained by differences in what the LLM
may `think' of as an appropriate level of phonetic transcription, and what the
creators of the Wiktionary data for the given language may have thought. For
example, noted above in Section~\ref{subsubsec:phonotactics}, the Wiktionary
entries for Japanese have exceedingly fine-grained phonetic transcriptions, so
that the word \emph{raigetsu} (`next month') is transcribed as
`\textfishhookr~\textsubbar{a}~i~g~\textsubbar{e}~\texttoptiebar{ts}~\textbari'
in Wiktionary, but would be transcribed much more phonemically in the system
generated by the LLM as `\textfishhookr~a~i~.~g~e~t~u'.

\subsection{Stories and grammatically targeted texts}
\label{subsec:stories}

The morphosyntactic component works by
transforming English text into the morphosyntactic
structure of the target language. For example, as
we document in more detail later on, for a
language that has singular-dual-plural marking on
nouns, the system would annotate SING, DUAL or
PLUR on nouns according to whether in the English
source sentence has a noun in the singular, the
plural but with the meaning of two, or plural with
a meaning of more than two.\footnote{There are
clearly pitfalls to this approach, the most
obvious being that terms in one language often do
not map one-to-one in their meanings or usages in
another. To take an example germane to the current
discussion, \emph{scissors} is a plural noun in
English (also in Spanish---\emph{tijeras}), but in
German it is singular (\emph{Schere}).}
For this purpose we need English texts from which to start, which we generated
in an initial data-creation phase.

First, using Claude 3.5 Sonnet, we generated 16 stories, using
the story generation prompt presented in
Appendix~\ref{subsubapp:story_prompt}. One example
of such a story is the following which, among
other things, contains an example of a noun
\emph{tower}, which should be rendered as DUAL in
a language that marks nouns with DUAL:

\begin{lstlisting}[language=bash]
The Two Towers

In a small town, there were two tall towers. One
was old and made of stone. The other was new and
made of glass. The people in the town liked to
talk about which tower was better.

Tom was a young boy who lived in the town. He
loved to look at both towers. Every day after
school, he would sit in the park between the
towers and draw pictures of them.

One day, a group of kids came up to Tom. They
asked him which tower he liked more. Tom thought
for a moment and said, "I like both
towers. They're different, but both are special."

The kids laughed at Tom. They said he had to pick
one. Some kids liked the old stone tower because
it was strong and had been there for a long
time. Other kids liked the new glass tower because
it was shiny and modern.

Tom felt sad. He didn't want to choose. He thought
both towers were important to the town.

That night, there was a big storm. The wind blew
hard, and rain poured down. In the morning, the
town woke up to see that both towers had been
damaged. The old stone tower had cracks, and some
of the glass on the new tower was broken.

The town mayor called a meeting. Everyone argued
about which tower to fix first. They didn't have
enough money to repair both at the same time.

Tom raised his hand and spoke up. He said, "Why
don't we work together to fix both towers? The old
tower is our history, and the new tower is our
future. We need both."

The people in the town thought about what Tom
said. They realized he was right. The two towers
were both important parts of their town.

So, the townspeople worked together. They fixed
the old stone tower and the new glass tower. It
took a long time, but they did it as a team.

When the work was done, the two towers stood tall
again. This time, they had something new. Between
them was a bridge that connected the towers. Tom's
idea had brought the town together.

From that day on, people stopped arguing about
which tower was better. They understood that both
towers made their town special. And Tom? He still
sat in the park, drawing pictures of the two
towers, now joined by a bridge of friendship.
\end{lstlisting}

In addition to these stories, we generated sets of
sentences that highlight particular grammatical
features. See subsequent prompts in
Appendix~\ref{subsubapp:story_prompt}. For
example, the following set of sentences gives
examples that highlight the first-person-plural
inclusive/exclusive distinction, which is of
course not marked morphologically in English, but
is in many languages:

\begin{lstlisting}[language=bash]
1. We should all pitch in to clean up the park. (INCLUSIVE)
2. We're going to the movies later, want to join us? (INCLUSIVE)
3. We've decided to go with a different contractor for this project. (EXCLUSIVE)
4. Let's work together to solve this problem, shall we? (INCLUSIVE)
5. We're not happy with your performance lately. (EXCLUSIVE)
6. We need to be more careful about recycling in our office. (INCLUSIVE)
7. We've already eaten dinner, but thanks for the invitation. (EXCLUSIVE)
\end{lstlisting}

\subsection{Lexicon creation}
\label{subsec:lexicon}

After the morphosyntactic modules have run the result are  files
containing the source sentences and the morphosyntactic annotations. The
phonotactic model is then used by the \texttt{lexicon.py} module to create
lexical entries for each of the morphemes, including stem forms and the various
morphosyntactic affixes created during the preceding phase.  In the current
version, stems have a minimum length in terms of numbers of phonemes (defaulting
to 5) and affixes a maximum length, currently set to 3.

Once the lexicon has been created containing all forms, the phonemic
transcription of each sentences is added, and the result saved out as a
JSON~Lines file.

\subsection{Orthography}
\label{subsec:orthography}

While most languages throughout history have been unwritten, it is fairly
standard for ConLangs to have a written form. In many cases this involves
inventing a new script, such as the Elvish Scripts from the Lord of the Rings
stories. The writing system might be segmental (as with Elvish) or some sort of
morphosyllabic or mixed logographic-phonetic system, like Chinese or
Japanese.
Inventing new scripts, and proposing logographic scripts both involve some
complexity that make them interesting research projects in and of themselves. To
create a new script requires a generative image model tuned for creating
script-like output, rather than general images.\footnote{There is an additional
issue, namely that once the script is created, ideally one would like to develop
an encoding system for it so that one can represent it in Unicode, and fonts so
that one can render text in it.}  For mixed-logographic scripts, one has to
define a manageable set of symbols---on the order of a few hundred---that can
used singly, or in combination, to represent morphemes, or some non-phonological
aspect of the morpheme, such as its meaning.

Our current approach is more modest, namely to have the system create a
segmental writing system using a known script and to write a Python program that
can convert from IPA transcriptions into the written form. The prompt we use for
this is given in Appendix~\ref{subsubapp:orthography_prompt}. This task still
requires some creativity on the part of the LLM, since there are in principle
many ways in which a script could be used to represent phonemes in a
language. The following listing is an example output where the system (in this
case using Claude~3.5~Sonnet) was
instructed to produce a Cyrillic-based orthography for a language with
Welsh-like phonotactics. Among the interesting choices are the spelling of the
voiceless lateral fricative /\textbeltl/ as
\textlangle\includegraphics[width=2mm]{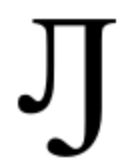}\textrangle. This is not part of
the core
Cyrillic alphabet, as used for Slavic languages such as Russian or Bulgarian: it
is an extension that was invented to write the same sound in Chukchi
(Chukotko–Kamchatkan language
family).\footnote{\url{https://en.wikipedia.org/wiki/Chukchi_language}.}

A sample listing is given below:

\includegraphics[width=\columnwidth]{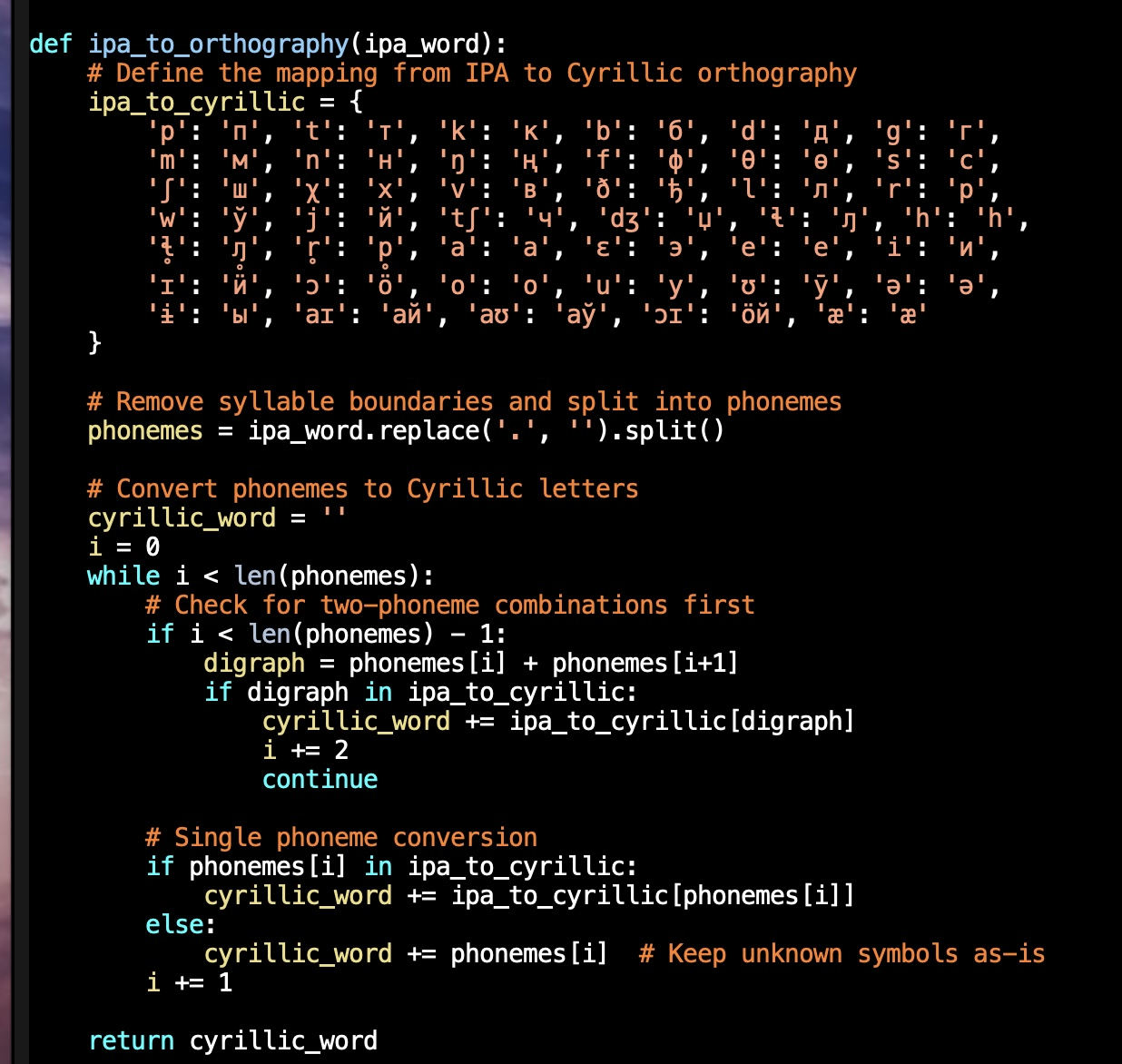}

\begin{figure*}
\includegraphics[width=\textwidth]{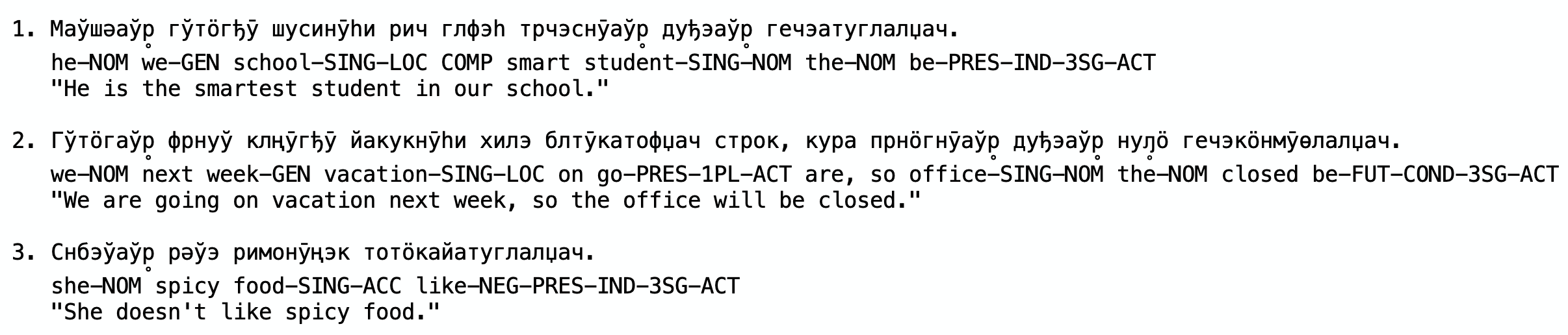}
\caption{\label{fig:welsh_cyrillic_sample}A sample of a language with Welsh-like
phonotactics and a Cyrillic-based orthography.}
\end{figure*}

A sample text and transcription are given in
Figure~\ref{fig:welsh_cyrillic_sample}.

Different LLMs exhibited different choices when it came to spelling given
phonemes. For example for the lateral fricative /ɬ/, with Latin script, Claude
3.5 Sonnet consistently chose \written{lh} as did
Qwen2.5-72B-Instruct. GPT-4o-mini also mostly chose this though in one instance
it chose \written{hl}. Gemini~2.5~Pro consistently chose \written{ll}, following
Welsh orthographic convention. When rendering in the Cyrillic script, Claude as
noted above chose
\textlangle\includegraphics[width=2mm]{Figures/chukchi_ll.png}\textrangle.
GPT-4o-mini on the other hand represented it with the IPA symbol \written{ɬ}.
Qwen2.5-72B-Instruct mostly chose the Cyrillic equivalent of \written{l} with a
circle under it indicating voicelessness, though in one instance it chose the
plain Cyrillic equivalent of \written{l}. Gemini~2.5~Pro chose a different
Cyrillic letter,
\textlangle\includegraphics[width=2mm]{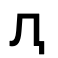}\textrangle, which
is used for
Khanty.\footnote{\url{https://en.wikipedia.org/wiki/El_with_descender}} Both
Gemini and Claude showed greater
`creativity' in finding a representation for the /ɬ/ in Cyrillic than did the
other two LLMs.

With the orthographic rules in place, the \texttt{lexicon.py} module used in
creating the phonological forms of the words (Section~\ref{subsec:lexicon}) is
run again, this time converting the phonemic transcriptions into orthographic
renditions. These are added to the lexicon as spellings for the stems and
affixes, and the corpus of sentences updated with the written forms of the
sentences.

One outstanding question is how one evaluates the orthography. The rules
generated by the system are consistent in that it produces one-for-one mappings
between phonemes and graphemes, leading to a very shallow orthography. But are
the mappings reasonable? As noted above, Claude Sonnet 3.5
`discovered' the use of `\includegraphics[width=2mm]{Figures/chukchi_ll.png}'
for /\textbeltl/, which is a perfectly defensible choice. But real orthographies
can often make seemingly odd choices. Some of these, such as the use of \written{gh} in
English to represent some instances of post-vocalic /f/, are due to historical
change. In other cases the choice was made more recently and cannot obviously be
explained in that way. For example, the use of \written{xh} in Albanian to represent
/\textyogh/ is a somewhat odd choice when compared against the use of /x/ in
most other languages that use the Latin script.  While not an orthography per
se, but rather a pedagogical aid, the Mandarin phonetic alphabet Hanyu Pinyin
for the most part made conventional choices for representing Mandarin
segments. Nonetheless, its choice of \written{q} to represent /t{\textctc}\textsuperscript{h}/
is odd.  It is important to bear in mind that the use of a script in a
language's writing system involves conventional choices for how each symbol in
the script is used. By and large, writing systems tend to obey these
conventions, but there is no hard and fast rule that they must do so.

\subsection{Handbook generation}
\label{subsec:handbook_generation}

The final stage involves creating a handbook, basically a short grammatical
description of the created language. The template prompt for this stage is
given in Appendix~\ref{subsubapp:handbook_prompt}.  The LLM is given the
phonology, breaking down the phonemes into consonants and vowels, as is standard
practice in grammar writing. Next the model is given the code that generates the
orthography. A sample of the lexicon is presented, and morphosyntactic settings
along with reminders about what these mean. A short set of sample texts are then
presented. The specification of the format of the handbook is given last.

Note that the very first instruction given is to create a name for the
language. As we see in the next section, there are some interesting biases
exhibited by the LLM in its choice of names.  An example handbook for
``Kethébik'', an ergative-absolutive language with Turkish-like features based
on Welsh phonotactics and using the Roman script is given below:

\begin{promptbox}{Kethébik handbook}{Examples/kethebik.txt}
  \end{promptbox}

\subsubsection{Biases in language name choice}
\label{subsubsec:name_biases}

As noted above, when creating the handbook, the system was instructed to create
a name for the ConLang. Interestingly, though ultimately not surprisingly, the
system's choices were heavily biased by the script chosen for the language's
orthography. See Table~\ref{tab:language_names}. The tendencies seem clearest
for Claude, but can be found for other models too.

The data in Table~\ref{tab:language_names} were derived by creating 117
handbooks using Gemini-2.6-Pro, 126 using Claude 3.5 Sonnet, 122 using
Qwen2.5-72B-Instruct and 125 using GPT-4o-mini. The reason for the difference in
numbers relates to the fact that some
models are better at others at creating working code to generate the
orthographies. In this case we ran the handbooks with 4 base
phonotactics---Japanese, Welsh, French, Spanish; 4 morphosyntactic
bundles---turkish, arabic, mizo and hixkaryana; and 4 scripts---Cyrillic,
Arabic, Latin, Greek. All of these were generated using Claude. Then for each of
the 64 combinations, the four LLMs were used to create an orthography and a
handbook. Since some of the LLMs failed in several cases, either at the
orthography phase, or at the handbook generation phase, we reran a second time,
this time using Claude's orthographies, and just using the four LLMs to create
the handbooks.

When the Latin script was chosen, the names chosen had a fantasy feel to
them---the kind of name one might associate with mythical races in fantasy
novels: \emph{Seran}, \emph{Althar}, \emph{Sérulang}, \emph{Nani}.
For Cyrillic, the names look decidely more `Slavic' or at least Central Asian,
though sometimes `boring' names like \emph{Cyrillic Ergative}, \emph{Cyrillic} or variants
like \emph{Kyrillic} are chosen.
For Arabic script, in contrast, the names sound a
bit Afghan or Persian.
Finally when the Greek script was chosen, the names
were all formed from Greek roots or at least roots that look Greek. Note that
the names chosen by Qwen were actually presented in Greek script, but have been
transliterated in the table.

\begin{table*}[t]
\begin{center}
{\tiny
\begin{tabular}{|l|lr|lr|lr|lr|}
\dtoprule
 & \multicolumn{2}{|c|}{Gemini-2.5-Pro} & \multicolumn{2}{|c|}{Claude 3.5 Sonnet} & \multicolumn{2}{|c|}{Qwen2.5-72B-Instruct} & \multicolumn{2}{|c|}{GPT-4o-mini}\\
 & Name & Count & Name & Count & Name & Count & Name & Count\\
\hline
Latin & Seran & 3/30 (0.10) & Zohulang & 2/32 (0.06) & Althar & 3/28 (0.11) & Zaru & 2/32 (0.06)\\
 & Véronien & 2/30 (0.07) & Zashuni & 2/32 (0.06) & Alorin & 3/28 (0.11) & Yonu & 2/32 (0.06)\\
 & Sotari & 2/30 (0.07) & Thrlangai & 2/32 (0.06) & Lirin & 2/28 (0.07) & Nani & 2/32 (0.06)\\
 & Zanari & 1/30 (0.03) & Sérulang & 2/32 (0.06) & Érhláth & 1/28 (0.04) & Frchesh & 2/32 (0.06)\\
 & Véronel & 1/30 (0.03) & Sénalang & 2/32 (0.06) & Élvarin & 1/28 (0.04) & Zohari & 1/32 (0.03)\\
\hline
Greek & Kallisti & 9/30 (0.30) & Zethalian & 2/30 (0.07) & Kyrinian & 5/31 (0.16) & Lurian & 2/31 (0.06)\\
 & Aethelian & 6/30 (0.20) & Zetakos & 2/30 (0.07) & Synthetike  & 1/31 (0.03) & Kɨrɨn & 2/31 (0.06)\\
 & Korythian & 4/30 (0.13) & Neosyllabic & 2/30 (0.07) & Skylan  & 1/31 (0.03) & Kysar & 2/31 (0.06)\\
 & Zorani & 1/30 (0.03) & Neolexia & 2/30 (0.07) & Phrantik  & 1/31 (0.03) & Koinos & 2/31 (0.06)\\
 & Rhovanic & 1/30 (0.03) & Neolekta & 2/30 (0.07) & Phragmatiko  & 1/31 (0.03) & Zmthif & 1/31 (0.03)\\
\hline
Arabic & Zarani & 5/28 (0.18) & Thulāni & 2/32 (0.06) & Alathian & 5/31 (0.16) & Koshari & 3/31 (0.10)\\
 & Zarish & 3/28 (0.11) & Thraenian & 2/32 (0.06) & Alarabiya & 5/31 (0.16) & Ghrin & 2/31 (0.06)\\
 & Zarathian & 2/28 (0.07) & Sitara & 2/32 (0.06) & Aramith & 2/31 (0.06) & Zukari & 1/31 (0.03)\\
 & Kaskari & 2/28 (0.07) & Shalāni & 2/32 (0.06) & Aralish & 2/31 (0.06) & Zohor & 1/31 (0.03)\\
 & Zulkhari & 1/28 (0.04) & Shalāmik & 2/32 (0.06) & Aralang & 2/31 (0.06) & Zaru & 1/31 (0.03)\\
\hline
Cyrillic & Vesnyan & 3/29 (0.10) & Cyrillic Ergative  & 6/32 (0.19) & Kyrillic & 9/32 (0.28) & Zohun & 2/31 (0.06)\\
 & Kherzovan & 2/29 (0.07) & Cyrillic & 5/32 (0.16) & Varnan & 5/32 (0.16) & Zaru & 2/31 (0.06)\\
 & Zoryanin & 1/29 (0.03) & Krylovik & 3/32 (0.09) & Krylthar & 5/32 (0.16) & Lurian & 2/31 (0.06)\\
 & Zhalian  & 1/29 (0.03) & Vornese & 2/32 (0.06) & Kryslan & 3/32 (0.09) & Lodi & 2/31 (0.06)\\
 & Voryenian & 1/29 (0.03) & Koshanese & 2/32 (0.06) & Vrylak & 1/32 (0.03) & Kɨrɨn & 2/31 (0.06)\\
\dbottomrule
\end{tabular}
} 
\caption{\label{tab:language_names}Top five language names chosen by each LLM for  each of the four scripts..}
\end{center}
\end{table*}

These choices clearly reflect the biases that are present in the data that the
LLM will have been exposed to.  The Latin script is the most widely used script
in the world today, being used to write languages on five continents. It is
therefore hard to localize it to any particular area. Presumably the choice of
fantasy-sounding names reflects an association between ConLangs and the world of
fantasy fiction.

Cyrillic on the other hand, is much more localized, being used to write
languages of Eastern Europe and Central Asia, in particular for languages spoken
in the area of the former Soviet Union. Thus names like \emph{Krylovik}
which sounds decidedly Slavic, or \emph{Kherzovan}, which sounds `Central Asian'.

Arabic is somewhat less localized than Cyrillic historically, having been used
historically to write the languages of Muslim populations in many parts of the
world---see the introduction of \citet{Doctor:EtAl:22} for some discussion---but
today is largely associated with Western Asia and North Africa, and is
particularly associated with major languages like Arabic, Persian, Urdu, Panjabi
and Sindhi. Hence names like \emph{Shadhiri} or \emph{Khorasani}.

Finally, Greek is the exception among these four scripts in that while there is
no principled reason it could not have been adapted to other
languages,\footnote{And indeed it was the parent of multiple other scripts,
including Latin and Cyrillic.} in fact
it is \emph{only} used to write Greek.  Hence the bias towards distinctly
Greek-sounding names.

This conflation of script and language is perhaps unfortunate, but at the same
time it reflects a similar conflation that even professional computational
linguists frequently make, as documented in \citet{Gorman:Sproat:22}.

\subsection{Translation of new texts}
\label{subsec:translation}

Based on the grammatical information created in the morphosyntactic component,
the handbook, and a sample of interlinear gloss translations, the system can
then be prompted to provide translations for new texts. The template prompt for
this stage is given in Appendix~\ref{subsubapp:translation_prompt}. As before,
the ``translations'' are really interlinear glosses, and after this translation
is performed, the lexicon creation module described in
Sections~\ref{subsec:lexicon} and \ref{subsec:orthography} are called to augment
the lexicon as needed with new entries and their orthographic forms.

A couple of example translations for a language called Zashuni based on arabic
(VSO order), with dual number marking, using Japanese-like phonotactics and
Latin orthography is given below. The LLM used was Claude 3.5 Sonnet.  Note the
correct translation of \emph{two other chicks} into dual number marking, as well
as \emph{both towers} into the dual, even though \emph{both} had not occurred in
the original set of example sentences used to create the morphosyntax. However
the use of the plural in the following sentence on the adjective
\emph{different}, which probably should be dual, suggests that there is room for
improvement.

\begin{promptbox}{Zashuni translation}{Examples/spring.txt}
  \end{promptbox}


\subsection{Applications to Low-Resource Languages}
\label{subsec:low_resource}

LLMs' performance on linguistic tasks depends on their being exposed to large
amounts of training data, but this immediately brings up the question of what
these models can do to help with low-resource languages which, by definition,
have very little in the way of training data.

One piece of work which attracted a lot of attetion was the \emph{Machine
Translation from One Book} project \citep{Tanzer:EtAl:24}, which purported to
build a machine translation system by presenting a single grammar book of a
low-resource language, the Trans-New-Guineau language Kalamang, as part of the
prompt to a version of Gemini. In principle the LLM might have `read' and
`understood' the grammar presentation, and used that knowledge, along with the
examples presented in the book, to construct an understanding of Kalamang
grammar, which it could then use to translate novel
examples. \citet{Aycock:EtAl:25} argue that, in fact, the system very likely
picked up whatever it learned about Kalamang from the examples, rather than from
any understanding of the grammatical description.

But no matter. It is still an interesting question to see to what extent LLMs
can be used to help with low-resource languages. Such a project falls within the
scope of what one might call the ``AI Linguist'', which is modeled on the term ``AI
Scientist'' \citep{Lu:EtAl:2024}. The AI Linguist is an artificial system
capable of behaving like a trained field linguist, able to assemble disparate
pieces of information about a language, and synthesize them into an
understanding of how that language works. An example of such an approach is that
of \citet{Torrent:EtAl:24}, who use LLMs as a linguist's aid in research in
Construction Grammar.

One possible application of the ConLang approach we have presented here is to
improving translation from high-resource languages into low-resource
languages. As long as the LLM has been exposed to enough data from a
low-resource language, translation from that language into a high-resource
language can often produce reasonable results. This is simply because the LLM
has a good model of the target language and so can construct sensible
translations given the meanings of the words in the source language. Going in
the other direction is another matter, since the LLM will not typically have a
very good model of the target language.

One possible approach to improving translation into low-resource languages is to
distill knowledge about the language into a set of transformations that can
modify the high-resource source into a structure similar to that of the target
langauge. For example if translating the English sentence ``John ate cucumbers''
into an SOV Ergative-Absolutive language with overt case suffixes, one might
first transform the sentence into something like ``John-ERG cucumber-PL-ABS
eat-PAST''. Approaches along these lines are not new: \citet{Ray:25} discusses
one of the first ``mechanical translation'' systems by \cite{Pucci:31}, which
involved annotating grammatical information on source sentences to aid in the
translation to the target language.  But it is also a natural extension of the
approach to ConLangs presented here.

We present here a simple experiment along these lines for translating from
English into Ainu, a severely endangered language isolate of Hokkaido,
Japan. While the results are largely negative, we present some evidence that
with improvements to the ConLang morphosyntactic approach we have presented
here, one could achieve actual gains.

The basic idea is to improve translation from English into Ainu, by annotating
the English source sentence to look more like the target language, using the
approach developed in Section~\ref{sec:morphosyntax}.  The first task is to
create a set of morphosyntactic specifications appropriate for Ainu.  For this
we used NotebookLM (\url{https://notebooklm.google.com/}) to summarize important
Ainu inflection affixes from the grammatical description in
\citet{Nakagawa:24}.\footnote{We thank Hiroshi Nakagawa for kindly allowing us
to use the PDF of his grammar for research purposes.}
The resulting table was then fed to Claude Code
(\url{https://docs.claude.com/en/docs/claude-code/sdk/sdk-overview}), along with
our morphosyntax specifications described earlier, with an instruction to create
a similar specification tailored to Ainu. The result was then hand-checked and
modified in cases where the automatically generated specifications deviated from
our own understanding of Ainu grammar.

For translation material we chose 50 example sentences from the Ainu textbook by
\citet{Nakagawa:13}, where we translated the original Japanese sentence into
English. Note that there are also online Ainu lessons in English such as
\url{https://unilang.org/course.php?res=58&subid=1}, which provide reasonable
translation pairs, but given our experience with using these online materials,
we suspect data leakage.

We used two LLMs, Claude Sonnet 3.5 and Gemini 2.5 Pro to annotate the English
sentences using the morphosyntactic component of our system. In addition to
these two annotations, we also prepared by hand a very literal back translation
from Ainu into English of the target Ainu sentence. Our purpose in doing this is
to test the limits of the proposed approach by providing a version of
English sentence that is close to an ideal modification towards the
target-language structure. An example of an Ainu sentence, its English
translation, and the three modifications of the English sentence is given
in Table~\ref{tab:ainu_exx}. Note that both Claude and Gemini render rather sparse annotations,
with Claude in particular omitting the possessive/verbal second-singular
prefix. Ainu has relatively sparse inflection, and this result mirrors our
observation that the models generally
do better when there is more information that they need to mark.

\begin{table*}
\centering
\begin{tabular}{ll}
Ainu & makanak e=iki wa e=teke arka?\\
English & How did you hurt your hand?\\
Claude Sonnet 3.5 & you ACT-hurt your hand-SING how hurt-PAST?\\
Gemini 2.5 Pro & you your hand how 2-SG-hurt?\\
Hand-annotated & how 2SGNOM-do and 2SG-hand hurt?\\
\end{tabular}
\caption{\label{tab:ainu_exx}Ainu example sentence with translation and
  morphosyntactic glossing from various systems.}
\end{table*}

We then used Claude Sonnet 3.5 to translate from English to Ainu, under two
conditions: with the English source sentence alone, and the English source
sentence coupled with one of the annotations. We ran this five times for each of
the three annotation methods. Corpus bigram, trigram and tetragram BLEU scores are shown in
Table~\ref{tab:bleu}.
\begin{table*}[t]
\begin{center}
\begin{tabular}{llrrr}
\toprule
System & Condition & Bigram BLEU & Trigram BLEU & Tetragram BLEU\\
\midrule
Claude & Unannotated & \textbf{18.15} & \textbf{11.71} & \textbf{7.69}\\
Claude & Annotated & 15.46 & 10.06 & 5.87\\
\midrule
Gemini & Unannotated & \textbf{18.40} & \textbf{11.11} & \textbf{6.27}\\
Gemini & Annotated & 13.25 & 7.29 & 3.82\\
\midrule
Hand & Unannotated & 17.00 & 10.29 & 5.99\\
Hand & Annotated & \textbf{23.17} & \textbf{15.23} & \textbf{10.36}\\
\bottomrule
\end{tabular}
\caption{\label{tab:bleu}BLEU scores for each of the three annotation
  conditions, averaged over five runs.}
\end{center}
\end{table*}
As before, Claude performs better than Gemini on average, but translating with
the Claude annotation still yields lower BLEU scores than translating with the
English source alone. On the other hand, the hand annotated data shows a
dramatic reversal, with the annotation yielding a substantial improvement in BLEU
scores over just using the English source alone.

Clearly our current morphosyntactic annotation system is not yet adequate to the
task of aiding translation for low-resource languages.  But the results with the
hand annotation suggest that an ideal future annotation system could yield
measurable gains in low-resource language translation. An area of future research is
improvement of the morphosyntactic annotation: the current output ConLangs are
still too parasitic on English, so development of methods to make the
annotations less English-like is desirable in any case.

\subsection{Prompts}
\label{subapp:prompts}

\subsubsection{System Prompt}
\label{subsubapp:system_prompt}

We prefixed the user prompts below with the following:

\begin{quote}
You are an expert linguist who is also an expert Python programmer.  You know a
wide variety of languages, and your hobby is creating Constructed Languages
--- ConLangs.
\end{quote}

\subsubsection{Phonotactics Prompt}
\label{subsubapp:phonotactics_prompt}

\begin{promptbox}{Phonotactics prompt}{Prompts/phonotactics_prompt.txt}
\end{promptbox}

\subsubsection{Phonotactics Improvement Prompt}
\label{subsubapp:phonotactics_improvement_prompt}

The following is an example of a prompt generated for the second round of the
agentic phonotactic system for a Welsh-based phonotactics.

\begin{promptbox}{Phonotactics improvement prompt}{Prompts/phonotactics_improvement_prompt_full.txt}
\end{promptbox}

\subsubsection{Story Generation Prompts}
\label{subsubapp:story_prompt}

\begin{promptbox}{Story generation prompt}{Prompts/story_generation.txt}\end{promptbox}
\begin{promptbox}{Sentence design general guidelines}{Prompts/sentence_design/general_guidelines.txt}\end{promptbox}
\begin{promptbox}{Comparative examples}{Prompts/sentence_design/comparative.txt}\end{promptbox}
\begin{promptbox}{Core argument examples}{Prompts/sentence_design/core_arguments.txt}\end{promptbox}
\begin{promptbox}{Inclusive/exclusive examples}{Prompts/sentence_design/inclusive_exclusive.txt}\end{promptbox}
\begin{promptbox}{Negative examples}{Prompts/sentence_design/negative.txt}\end{promptbox}
\begin{promptbox}{Nominal number examples}{Prompts/sentence_design/nominal_number.txt}\end{promptbox}
\begin{promptbox}{Oblique argument examples}{Prompts/sentence_design/oblique_arguments.txt}\end{promptbox}
\begin{promptbox}{Person examples}{Prompts/sentence_design/person.txt}\end{promptbox}
\begin{promptbox}{Tense/aspect examples}{Prompts/sentence_design/tense_aspect.txt}\end{promptbox}
\begin{promptbox}{Voice examples}{Prompts/sentence_design/voice.txt}\end{promptbox}

\subsubsection{Morphosyntax Prompts}
\label{subsubapp:morphosyntax_prompts}

\begin{promptbox}
    {Cumulative morphosyntax prompt for grammatical case}
    {Prompts/cumulative_morphosyntax/case.txt}
\end{promptbox}

\subsubsection{Orthography Prompt}
\label{subsubapp:orthography_prompt}

\begin{promptbox}{Orthography prompt}{Prompts/orthography_prompt.txt}
\end{promptbox}

\subsubsection{Handbook Prompt}
\label{subsubapp:handbook_prompt}

\begin{promptbox}{Handbook prompt}{Prompts/handbook_prompt.txt}
\end{promptbox}

\subsubsection{Translation Prompt}
\label{subsubapp:translation_prompt}

\begin{promptbox}{Translation prompt}{Prompts/translation_prompt.txt}
\end{promptbox}



\subsection{Phonological changes}
\label{subapp:phonological_changes}

Besides phonotactics, all languages also exhibit phonological processes that may
alter the shapes of words and morphemes from what one would expect from the
underlying phonotactic model.  Historically such changes start out as
articulatory or acoustically motivated phonetic changes, which over time may
lose their immediate phonetic motivation and become part of the
morphophonological pattern of the language \citep{Blevins:04}. For example, the
\emph{umlaut} vowel change observed in the formation of some plural nouns in
German like \emph{Maus} `mouse' (plural \emph{M\underline{\"au}ser}),
\emph{Haus} `house' (plural \emph{H\underline{\"au}ser}), or \emph{Mann} `man'
(plural \emph{M\underline{\"a}nner}), originated from a Proto-West-Germanic
plural formation with the vowel /i/, which caused a fronting of the vowel of the
preceding noun stem. The fronting itself was a phonetically motivated case of
assimilation. The /i/ was subsequently lost leaving only the vowel change in the
stem. This vowel change, now dissociated from the original phonetic motivation,
came itself to be associated with plural formation, and in fact spread in its
use to nouns that did not originally have the /i/ marking in the plural.

There have been hundreds of phonological changes documented for the world's
languages. Many of these can be grouped into larger categories, including:
\begin{description}
\item[Assimilation:] A sound becomes more similar to one in its neighborhood.
  E.g. the Germanic umlaut example given above.
\item[Dissimilation:] A sound becomes less similar to one in its neighborhood.
  E.g. French \emph{pe\underline{l}egrin} `pilgrim', from Latin
  \emph{pe\underline{r}egrinus}, with a change of the first /r/ to /l/,
  dissimilating from the second /r/.
\item[Deletion:] The removal of a sound. E.g. the loss of final consonants in
  French, such as \emph{chat} /{\textesh}a/ `cat' where the original final /t/,
  still spelled, is lost in the phonological form.
\item[Epenthesis:] The insertion of a sound. E.g. in Spanish, initial consonant
  clusters with /s/ are prepended by an /e/, thus \emph{\underline{e}staci\'on}
  `station'.
\item[Lenition:] Or weaking. E.g. Intervocalic /t/ and /d/ in most dialects of
  American English are weakened to a \emph{flap} sound so that \emph{butter} is
  /b{\textturnv}{\textfishhookr}{\textrhookschwa}/.
\item[Fortition:] Or strengthening. E.g. Breton \emph{hard mutation} where
  initial voiced stops are changed to voiceless stops, as in \emph{breur}
  `brother' \emph{ho \underline{p}reur} `your (pl.) brother'.
\item[Sound shifts:]  Such as the Great English Vowel Shift that shifted most
  long vowels, with high vowels becoming
  diphthongs---/\=\i/$\rightarrow$/a\textsc{i}/,
  /\=u/$\rightarrow$/a\textupsilon/, mid vowels becoming high
  vowel (also pronounced as diphthongs in most varieties of English)---/\=o/$\rightarrow$/\=u/, /\=e/$\rightarrow$/\=\i/, and low vowels
  becoming mid vowels---/\=a/$\rightarrow$/\=e/.
\end{description}
Within each of these broad categories are many variations depending in part on
the phonotactic pattern of the language to which the changes were applied.

A main problem for any ConLang creator is thus to decide on a reasonable set of
phonological rules to apply. There has in fact been a lot of interest in this
problem in the ConLang community with various advice on how to `age' one's
ConLang, i.e. how to make it change over time in a way that resembles
phonological changes that have been attested in the world's languages.

One resource for this is
Diachronica\footnote{\url{https://chridd.nfshost.com/diachronica/index-diachronica.pdf}},
which compiles phoneme inventories for a large number of historically
reconstructed languages, and lists phonological changes that historical
linguists have proposed to derive daughter language forms.

In this work we propose a novel approach that builds on Diachronica and guides
an LLM to construct a set of phonological rules that can plausibly apply to a
language with the specified phonotactics.  Basically the process involves
finding a phoneme inventory of a language X in the Diachronica data that is most
similar to the phonotactics of our language, and then selecting some rules based
on the rules listed for X as the basis of our phonological rule set.

The steps can be summarized as follows:

\begin{itemize}
\item First, Diachronica only provides phoneme inventories for the initial
  reconstructed proto-language. For example, the document lists a reconstructed
  inventory for Proto-Indo-European, and rules to go from Proto-Indo-European to
  Common Germanic, from Common Germanic to West Germanic, from West-Germanic to
  Anglo-Frisian, from Anglo-Frisian to Old English, etc. But phoneme inventories
  are not directly provided for Common Germanic, West Germanic and so forth. In
  other words for a chain of rule sets
  $A{\rightarrow}B{\rightarrow}C{\rightarrow}D$\ldots, we have the reconstructed
  phoneme inventory for $A$, but not for $B$, $C$, $D$. In order to consider the
  reconstructed rules for, say, $C{\rightarrow}D$ as a model for our ConLang, we
  therefore need to have a phoneme inventory for $C$. This can be constructed,
  approximately, by applying $A{\rightarrow}B$ to the inventory in $A$, deducing
  the likely phoneme set for $B$, and then repeating the process. So the first,
  offline task for the LLM was to fill out the dataset provided by Diachronica
  by considering the phonological rules in each set, and deducing the phoneme
  sets of the corresponding daughter language.
\item
  Armed with the enhanced version of the Diachronica data, we then find the
  closest $N$ phoneme inventories to the one already developed for our ConLang.
\item
  The LLM is then presented with the phoneme inventory already created for the
  ConLang, a list of sample word forms generated by the phonotactic grammar, and
  for each of the $N$ phoneme inventories from Diachronica, the name of the
  reconstructed language, the reconstructed phoneme inventory, and the
  reconstruted phonological rules. The LLM is then instructed to look at the
  rules, and write a Python program that implements a reasonable subset of the
  rules that can be applied to the output of the phonotactic grammar. As with
  the phonotactic construction, the system is agentic in that it iterates the
  process, presenting the previous program and its outputs, and any issues
  noted, instructing the LLM to try to improve the implementation.
\end{itemize}

We describe each of these steps in more detail in the sections below.

\subsubsection{Enhancing Diachronica}
\label{subsubapp:enhancing_diachronica}

The first stage involves expanding Diachronica to provide phoneme inventories
for intermediate stage reconstructed languages. Starting with the proto language
for each major branch, rules to derive the daughter language are consulted, and
the phoneme inventory for the daughter language is estimated. The procedure is
then iterated, with the new daughter language and subsequent rules used to
derive the next level daughter languages.  At each stage the LLM is prompted to
provide its initial guess as to the phoneme inventory for the daughter language
using the prompt template Diachronica~Expansion given in
Appendix~\ref{subapp:phonology_prompts}. In a second stage, the LLM is asked
to possibly refine its estimate: see the prompt template in
Diachronica~Expansion~Stage~2.

For example the following phoneme set was reconstructed for Old English. Note
that this is constructed from Proto-Indo-European to Common Germanic to West
Germanic to Anglo-Frisian to Old English:

\begin{quote}
a, a{\textlengthmark}, b, d, d\textyogh, e, e{\textlengthmark}, f, g, i,
i{\textlengthmark}, j, k, k\textsuperscript{w}, l, m, n, o, o{\textlengthmark},
p, r, s, t, t\textesh, u, u{\textlengthmark}, w, x, \ae, \ae{\textlengthmark},
\~{o}{\textlengthmark}, \~{\i}{\textlengthmark}, \~{u}{\textlengthmark},
\textctj, \texttheta, \~{e}{\textlengthmark}.
\end{quote}

In the second version, Claude 3.5 Sonnet made the following decisions, in some cases
revising the inventory:
\begin{enumerate}
\item
Removed nasal vowels (\~{e}\textlengthmark, \~{o}\textlengthmark,
\~{\i}\textlengthmark, \~{u}\textlengthmark): The rule
``V[+nas]~$\rightarrow$~V[-nas]'' indicates that all nasal vowels became
non-nasal.
\item
Retained `h': This phoneme was present in Anglo-Frisian and there's no rule
indicating its loss, so it should be kept in Old English.
\item
Removed '\~{\textscripta}': The rule
``\~{\textscripta}\textlengthmark~$\rightarrow$~\~{o}\textlengthmark'' suggests
this phoneme changed, and then the nasalization was lost as per the second rule.
\item
Retained 'k': While there's a rule about `k' becoming `t\textesh' in certain
circumstances, it's not a blanket change. The original `k' should still exist in
other contexts.
\item
Added `\textctj': This comes from the rule
``k~\textgamma~g~$\rightarrow$~t\textesh~\textctj~d\textyogh'' although it's
noted to occur only in certain circumstances.
\item
Retained `u' and `i': The rule about their loss is context-dependent (word-final
position), so they should still exist in other positions.
\end{enumerate}
The second revised version was as follows:
\begin{quote}
a, a{\textlengthmark}, b, d, d\textyogh, e, e{\textlengthmark}, f, g, h, i,
i{\textlengthmark}, j, k, k\textsuperscript{w}, l, m, n, o, o{\textlengthmark},
p, r, s, t, t\textesh, u, u{\textlengthmark}, w, x, \ae, \ae{\textlengthmark},
\textctj, \texttheta.
\end{quote}

According to
Wikipedia\footnote{\url{https://en.wikipedia.org/wiki/Old_English_phonology}},
the actual Old English phoneme inventory was as follows:
\begin{quote}
\textscripta, \textscripta\textlengthmark, b, d, e, e\textlengthmark, f, h, i,
i\textlengthmark, j, k, l, m, n, o, o\textlengthmark, p, r, s, \textesh, t,
t\textesh, u, u\textlengthmark, w, x, \ae, \ae\textlengthmark, \textesh,
\texttheta, \textgamma, y, y\textlengthmark.
\end{quote}
Notable differences from the Diachronica-derived reconstruction include:
\begin{itemize}
\item Cases like /d\textyogh/, which was an allophone of /t\textesh/, but which the
  reconstruction includes as a separate phoneme. Similarly /g/ was an allophone
  of /k/.
\item Decisions like whether to consider the /k\textsuperscript{w}/ of a word
  like \emph{cw\=en} `queen' as a separate phoneme, or as a sequence of /kw/.
\item Missing /\textgamma/ and /\textesh/.
\item Use of /a/ rather than /\textscripta/.
\item The missing front rounded vowels /y/ and /y\textlengthmark/.
\end{itemize}
In addition, Old English had several diphthongs which were not listed as
separate phonemes in the Diachronica-derived reconstruction.

Despite these differences, the reconstruction is not too bad, considering that
it is derived from a hypothetical reconstruction of Proto-Indo-European, run
through four sets of imprecisely stated rules.

\subsubsection{Selecting the closest phoneme set}
\label{subsubapp:selecting_phoneme_set}

Given the expanded Diachronica set the next task is to find the language(s) with
the closest phoneme set to that of the constructed ConLang. For each phoneme set
from Diachronica, and for each phoneme in that set, the distance of the closest
phoneme in the ConLang phoneme set is found. Distance between two phonemes is
computed by counting the matching phonetic features of the two phonemes, using
the phonetic feature set from PHOIBLE~2.0 \citep{Phoible}. The distance between
the two phoneme sets is simply the sum of the phoneme distances, with an
additional penalty for the difference in the cardinalities of the two
sets. Diachronica phoneme sets are then rank ordered by increasing distance from
the ConLang set.

\subsubsection{Constructing the phonology}
\label{subsubapp:constructing the phonology}

Once the closest Diachronica-derived phoneme set is chosen, the next phase is to
come up with a set of phonological rules. These rules are based on the sets of
rules that are listed in Diachronica that mapped from the language associated
with the phoneme set, to various daughter languages.  Here we allow the LLM a
fair amount of freedom, basically telling it to ``Devise a set of rules that
makes sense for the phoneme set of our language given the observed
phonotactics'' and to implement this as an ordered set of rules in Python. The
prompt for this initial set is given under ``Phonological~Rules'' in
Appendix~\ref{subapp:phonology_prompts}. One important instruction is that as
part of the rule definition, the LLM must provide a set of example input-output
pairs that show what the model is supposed to do. These will then be used in
subsequent phases to check that the rules are doing what they are supposed to,
and to give feedback to the model to improve the rules.

An initial set of rules proposed for a Welsh-based phonology was as in the code
fragment below, where the closest phoneme set turned out to be Early Modern
English, and the basis rule sets chosen were Early Modern English to American
English, Early Modern English to Australian English and Early Modern English to
British English---see Sections~17.7.2.1.5 to 17.7.2.1.7 in Diachronica.

\begin{lstlisting}[language=Python,inputencoding=utf8]
import re

def vowel_shift(inp: str) -> str:
    # Example inputs/outputs:
    # u -> ʉ
    # o -> u
    # ɔ -> o
    vowel_map = {
        'u': 'ʉ',
        'o': 'u',
        'ɔ': 'o',
        'ɛ': 'e',
        'æ': 'ɛ'
    }
    return ' '.join(vowel_map.get(phone, phone) for phone in inp.split())

def consonant_lenition(inp: str) -> str:
    # Example inputs/outputs:
    # V t V -> V ð V
    # V k V -> V x V
    # V p V -> V f V
    inp = re.sub(r'([aɛeiɪɔouʊəɨæaɪaʊɔɪeɪəʊ]) t ([aɛeiɪɔouʊəɨæaɪaʊɔɪeɪəʊ])', r'\1 ð \2', inp)
    inp = re.sub(r'([aɛeiɪɔouʊəɨæaɪaʊɔɪeɪəʊ]) k ([aɛeiɪɔouʊəɨæaɪaʊɔɪeɪəʊ])', r'\1 x \2', inp)
    inp = re.sub(r'([aɛeiɪɔouʊəɨæaɪaʊɔɪeɪəʊ]) p ([aɛeiɪɔouʊəɨæaɪaʊɔɪeɪəʊ])', r'\1 f \2', inp)
    return inp

def nasal_assimilation(inp: str) -> str:
    # Example inputs/outputs:
    # n p -> m p
    # n k -> ŋ k
    # n . b -> m . b
    inp = re.sub(r'n (\. )?([pbm])', r'm \1\2', inp)
    inp = re.sub(r'n (\. )?([kg])', r'ŋ \1\2', inp)
    return inp

def palatalization(inp: str) -> str:
    # Example inputs/outputs:
    # t i -> tʃ i
    # d i -> dʒ i
    # s i -> ʃ i
    inp = re.sub(r't ([ieɪ])', r'tʃ \1', inp)
    inp = re.sub(r'd ([ieɪ])', r'dʒ \1', inp)
    inp = re.sub(r's ([ieɪ])', r'ʃ \1', inp)
    return inp

def vowel_nasalization(inp: str) -> str:
    # Example inputs/outputs:
    # a n -> ã n
    # o m -> õ m
    vowels = 'aɛeiɪɔouʊəɨæaɪaʊɔɪeɪəʊ'
    nasals = 'mnŋ'
    for v in vowels:
        for n in nasals:
            inp = re.sub(f'{v} ({n})', f'{ṽ} \1', inp)
    return inp

def cluster_simplification(inp: str) -> str:
    # Example inputs/outputs:
    # k t -> t
    # p t -> t
    # ŋ k -> ŋ
    inp = re.sub(r'[kp] t', 't', inp)
    inp = re.sub(r'ŋ k', 'ŋ', inp)
    return inp

def vowel_reduction(inp: str) -> str:
    # Example inputs/outputs:
    # a . k u -> ə . k u
    # k u . b r o -> k u . b r ə
    words = inp.split('#')
    for i, word in enumerate(words):
        syllables = word.strip().split('.')
        if len(syllables) > 1:
            for j, syl in enumerate(syllables[:-2]):  # All but the last two syllables
                syllables[j] = re.sub(r'([aɛeiɪɔouʊɨæaɪaʊɔɪeɪəʊ])', 'ə', syl)
        words[i] = '.'.join(syllables)
    return '#'.join(words)

def rules(inp: str) -> str:
    output = vowel_shift(inp)
    output = consonant_lenition(output)
    output = nasal_assimilation(output)
    output = palatalization(output)
    output = vowel_nasalization(output)
    output = cluster_simplification(output)
    output = vowel_reduction(output)
    return output
\end{lstlisting}

The rules are then run on words generated by the phonotactic model, as well as
the example input-output sequences listed in the phonological rule code itself,
and any errors noted. These errors, along with the original instructions and the
code produced on the previous phase, are fed to the model in a second round of
prompting.  For example, the code above misses spaces between phonemes in some
cases, so these and other issues are pointed out to the model. On the second
round the model is able to correct the spacing issue along with other problems,
listing its corrections as follows:

\begin{enumerate}
\item
Fixed the spacing issue by ensuring that all phonemes are properly
space-separated, including word and syllable boundaries.
\item
Corrected the `\verb|consonant_lenition|' function to properly apply the rules
for t~$\rightarrow$~\dh, k~$\rightarrow$~x, and p~$\rightarrow$~f between vowels.
\item
Fixed the `\verb|vowel_nasalization|' function to maintain the space between the
nasalized vowel and the following nasal consonant.
\item
Improved the `\verb|vowel_reduction|' function to correctly reduce vowels in
non-final syllables and the final syllable of words with more than one syllable.
\item
Updated the vowel set in various functions to include the new `\textbaru' phoneme
introduced by the vowel shift.
\end{enumerate}

The second round code is as follows:

\begin{lstlisting}[language=Python,inputencoding=utf8]
import re

def vowel_shift(inp: str) -> str:
    # Example inputs/outputs:
    # u -> ʉ
    # o -> u
    # ɔ -> o
    vowel_map = {
        'u': 'ʉ',
        'o': 'u',
        'ɔ': 'o',
        'ɛ': 'e',
        'æ': 'ɛ'
    }
    return ' '.join(vowel_map.get(phone, phone) for phone in inp.split())

def consonant_lenition(inp: str) -> str:
    # Example inputs/outputs:
    # V t V -> V ð V
    # V k V -> V x V
    # V p V -> V f V
    vowels = 'aɛeiɪɔouʊəɨæaɪaʊɔɪeɪəʊʉ'
    inp = re.sub(f'([{vowels}]) t ([{vowels}])', r'\1 ð \2', inp)
    inp = re.sub(f'([{vowels}]) k ([{vowels}])', r'\1 x \2', inp)
    inp = re.sub(f'([{vowels}]) p ([{vowels}])', r'\1 f \2', inp)
    return inp

def nasal_assimilation(inp: str) -> str:
    # Example inputs/outputs:
    # n p -> m p
    # n k -> ŋ k
    # n . b -> m . b
    inp = re.sub(r'n (\. )?([pbm])', r'm \1\2', inp)
    inp = re.sub(r'n (\. )?([kg])', r'ŋ \1\2', inp)
    return inp

def palatalization(inp: str) -> str:
    # Example inputs/outputs:
    # t i -> tʃ i
    # d i -> dʒ i
    # s i -> ʃ i
    inp = re.sub(r't ([ieɪ])', r'tʃ \1', inp)
    inp = re.sub(r'd ([ieɪ])', r'dʒ \1', inp)
    inp = re.sub(r's ([ieɪ])', r'ʃ \1', inp)
    return inp

def vowel_nasalization(inp: str) -> str:
    # Example inputs/outputs:
    # a n -> ã n
    # o m -> õ m
    vowels = 'aɛeiɪɔouʊəɨæaɪaʊɔɪeɪəʊʉ'
    nasals = 'mnŋ'
    for v in vowels:
        for n in nasals:
            inp = re.sub(f'({v}) ({n})', f'\\1̃ \\2', inp)
    return inp

def cluster_simplification(inp: str) -> str:
    # Example inputs/outputs:
    # k t -> t
    # p t -> t
    # ŋ k -> ŋ
    inp = re.sub(r'[kp] t', 't', inp)
    inp = re.sub(r'ŋ k', 'ŋ', inp)
    return inp

def vowel_reduction(inp: str) -> str:
    # Example inputs/outputs:
    # a . k u -> ə . k u
    # k u . b r o -> k u . b r ə
    words = inp.split('#')
    for i, word in enumerate(words):
        syllables = word.strip().split('.')
        if len(syllables) > 1:
            for j in range(len(syllables) - 2):  # All but the last two syllables
                syllables[j] = re.sub(r'([aɛeiɪɔouʊɨæaɪaʊɔɪeɪəʊʉ])', 'ə', syllables[j])
            syllables[-1] = re.sub(r'([aɛeiɪɔouʊɨæaɪaʊɔɪeɪəʊʉ])', 'ə', syllables[-1])  # Reduce final syllable
        words[i] = ' . '.join(syllables)
    return ' # '.join(words)

def rules(inp: str) -> str:
    output = vowel_shift(inp)
    output = consonant_lenition(output)
    output = nasal_assimilation(output)
    output = palatalization(output)
    output = vowel_nasalization(output)
    output = cluster_simplification(output)
    output = vowel_reduction(output)
    return output
\end{lstlisting}

\subsection{Phonology Prompts}
\label{subapp:phonology_prompts}

The next two prompt templates relate to the expansion of the Diachronica data as
described in Section~\ref{subsubapp:enhancing_diachronica}. Note that the
\texttt{abbreviations} variable is filled in with a set of abbreviations listed
in \url{https://chridd.nfshost.com/diachronica/index-diachronica.pdf}, page 6.

\begin{promptbox}{Diachronica Expansion}{Prompts/diachronica.txt}
\end{promptbox}

\begin{promptbox}{Diachronica Expansion Stage 2}{Prompts/diachronica_2.txt}
\end{promptbox}

\begin{promptbox}{Phonological Rules}{Prompts/phonology_prompt.txt}
\end{promptbox}

\section{Morphosyntactic feature specifications}\label{sec:morphosyntax-params}

Morphosyntactic feature specifications based on nine languages are presented in
Tables~\ref{tab:arabic-spec}--\ref{tab:hard-language-spec}.

\begin{table*}[t]
    \centering
    \scriptsize
    \setlength{\tabcolsep}{5pt}
    \begin{tabular}{@{}lllp{18em}@{}} \toprule
        Submodule & Features & Subfeatures & Value \\
        \midrule
        Syntax & \texttt{main\_word\_order} & & VSO \\
        & \texttt{oblique\_word\_order} & & VOX \\
        & \texttt{adj\_noun\_word\_order} & & NA \\
        & \texttt{posspron\_pron\_word\_order} & & NPoss \\
        & \texttt{num\_noun\_word\_order} & & NumN \\
        & \texttt{adposition\_noun\_word\_order} & & PN \\
        \midrule
        Morphology & \texttt{case} & \texttt{case\_marking} & nominative, accusative, genitive \\
        & & \texttt{case\_marking\_strategy} & suffix \\
        & & \texttt{oblique\_case\_marking} & genitive \\
        & \texttt{definiteness} & \texttt{definiteness} & definite \\
        & & \texttt{definiteness\_marking\_strategy} & prefix \\
        & & \texttt{definiteness\_agreement} & None \\
        & \texttt{adjective\_agreement} & \texttt{adjective\_agreement} & number, case, definiteness \\
        & & \texttt{adjective\_agreement\_strategy} & suffix \\
        & \texttt{comparative} & \texttt{comparative} & comparative, superlative \\
        & & \texttt{comparative\_marking\_strategy} & suffix \\
        & \texttt{tense\_aspect} & \texttt{tense\_aspect} & present, past, future \\
        & & \texttt{tense\_aspect\_marking\_strategy} & suffix \\
        & \texttt{mood} & \texttt{mood} & indicative, subjunctive, imperative \\
        & & \texttt{mood\_marking\_strategy} & suffix \\
        & \texttt{voice} & \texttt{voice} & active, passive \\
        & & \texttt{voice\_marking\_strategy} & suffix \\
        & \texttt{person} & \texttt{person\_agreement} & first, second, third \\
        & & \texttt{person\_marking\_strategy} & suffix \\
        & & \texttt{verbal\_number\_agreement} & singular, plural, dual \\
        & & \texttt{verbal\_number\_marking\_strategy} & suffix \\
        & \texttt{inclusive\_exclusive} & & True \\
        & \texttt{nominal\_number} & \texttt{nominal\_number} & singular, plural, dual \\
        & & \texttt{nominal\_number\_marking\_strategy} & suffix \\
        & \texttt{relativization} & \texttt{relativization\_order} & head-initial \\
        & & \texttt{relativization\_marking} & head-marking \\
        & & \texttt{relativizer\_position} & postpositional \\
        & & \texttt{relativizer\_morpheme} & word \\
        & \texttt{negation} & & prepositional word \\
        & \texttt{infinitive} & \texttt{infinitive} & None \\
        \bottomrule
    \end{tabular}
    \caption{The configuration of the Arabic-like feature set.
    Though the Arabic inflectional paradigms are highly fusional rather than agglutinative, such features are expressed as suffixes or prefixes in this feature set for simplicity.}
    \label{tab:arabic-spec}
\end{table*}

\begin{table*}[t]
    \centering
    \scriptsize
    \setlength{\tabcolsep}{5pt}
    \begin{tabular}{@{}lllp{18em}@{}} \toprule
        Submodule & Features & Subfeatures & Value \\
        \midrule
        Syntax & \texttt{main\_word\_order} & & VOS \\
        & \texttt{oblique\_word\_order} & & VOX \\
        & \texttt{adj\_noun\_word\_order} & & NA \\
        & \texttt{posspron\_pron\_word\_order} & & NPoss \\
        & \texttt{num\_noun\_word\_order} & & NumN \\
        & \texttt{adposition\_noun\_word\_order} & & PN \\
        \midrule
        Morphology & \texttt{case} & & None \\
        & \texttt{definiteness} & & None \\
        & \texttt{adjective\_agreement} & & None \\
        & \texttt{comparative} & \texttt{comparative} & comparative, superlative \\
        & & \texttt{comparative\_marking\_strategy} & postpositional word \\
        & \texttt{tense\_aspect} & & None \\
        & \texttt{mood} & & None \\
        & \texttt{voice} & & None \\
        & \texttt{person} & \texttt{person\_agreement} & first, second, third \\
        & & \texttt{person\_marking\_strategy} & prepositional word \\
        & & \texttt{verbal\_number\_agreement} & singular, plural, dual, paucal \\
        & & \texttt{verbal\_number\_marking\_strategy} & prepositional word \\
        & \texttt{inclusive\_exclusive} & & False \\
        & \texttt{nominal\_number} & & None \\
        & \texttt{relativization} & \texttt{relativization\_order} & head-initial \\
        & & \texttt{relativization\_marking} & None \\
        & & \texttt{relativizer\_position} & None \\
        & & \texttt{relativizer\_morpheme} & None \\
        & \texttt{negation} & & prepositional word \\
        & \texttt{infinitive} & \texttt{infinitive} & None \\
        \bottomrule
    \end{tabular}
    \caption{The configuration of the Fijian-like feature set.}
    \label{tab:fijian-spec}
\end{table*}

\begin{table*}[t]
    \centering
    \scriptsize
    \setlength{\tabcolsep}{5pt}
    \begin{tabular}{@{}lllp{18em}@{}} \toprule
        Submodule & Features & Subfeatures & Value \\
        \midrule
        Syntax & \texttt{main\_word\_order} & & SVO \\
        & \texttt{oblique\_word\_order} & & VOX \\
        & \texttt{adj\_noun\_word\_order} & & NA \\
        & \texttt{posspron\_pron\_word\_order} & & PossN \\
        & \texttt{num\_noun\_word\_order} & & NumN \\
        & \texttt{adposition\_noun\_word\_order} & & PN \\
        \midrule
        Morphology & \texttt{case} & & None \\
        & \texttt{definiteness} & \texttt{definiteness} & definite, indefinite \\
        & & \texttt{definiteness\_marking\_strategy} & prepositional word \\
        & & \texttt{definiteness\_agreement} & None \\
        & \texttt{adjective\_agreement} & \texttt{adjective\_agreement} & number \\
        & & \texttt{adjective\_agreement\_strategy} & suffix \\
        & \texttt{comparative} & \texttt{comparative} & comparative, superlative, equative \\
        & & \texttt{comparative\_marking\_strategy} & prepositional word \\
        & \texttt{tense\_aspect} & \texttt{tense\_aspect} & present, past, future, imperfect \\
        & & \texttt{tense\_aspect\_marking\_strategy} & suffix \\
        & \texttt{mood} & \texttt{mood} & indicative, subjunctive, imperative, conditional \\
        & & \texttt{mood\_marking\_strategy} & suffix \\
        & \texttt{voice} & \texttt{voice} & active, passive \\
        & & \texttt{voice\_marking\_strategy} & suffix \\
        & \texttt{person} & \texttt{person\_agreement} & first, second, third \\
        & & \texttt{person\_marking\_strategy} & suffix \\
        & & \texttt{verbal\_number\_agreement} & singular, plural \\
        & & \texttt{verbal\_number\_marking\_strategy} & suffix \\
        & \texttt{inclusive\_exclusive} & & False \\
        & \texttt{nominal\_number} & \texttt{nominal\_number} & singular, plural \\
        & & \texttt{nominal\_number\_marking\_strategy} & suffix \\
        & \texttt{relativization} & \texttt{relativization\_order} & head-initial \\
        & & \texttt{relativization\_marking} & head-marking \\
        & & \texttt{relativizer\_position} & postpositional \\
        & & \texttt{relativizer\_morpheme} & word \\
        & \texttt{negation} & & postpositional word \\
        & \texttt{infinitive} & \texttt{infinitive} & infinitive \\
        & & \texttt{infinitive\_position} & suffix \\
        \bottomrule
    \end{tabular}
    \caption{The configuration of the French-like feature set.}
    \label{tab:french-spec}
\end{table*}

\begin{table*}[t]
    \centering
    \scriptsize
    \setlength{\tabcolsep}{5pt}
    \begin{tabular}{@{}lllp{18em}@{}} \toprule
        Submodule & Features & Subfeatures & Value \\
        \midrule
        Syntax & \texttt{main\_word\_order} & & OVS \\
        & \texttt{oblique\_word\_order} & & OVX \\
        & \texttt{adj\_noun\_word\_order} & & NA \\
        & \texttt{posspron\_pron\_word\_order} & & PossN \\
        & \texttt{num\_noun\_word\_order} & & NumN \\
        & \texttt{adposition\_noun\_word\_order} & & PN \\
        \midrule
        Morphology & \texttt{case} & & None \\
        & \texttt{definiteness} & \texttt{definiteness} & definite, indefinite \\
        & & \texttt{definiteness\_marking\_strategy} & prepositional word \\
        & & \texttt{definiteness\_agreement} & None \\
        & \texttt{adjective\_agreement} & \texttt{adjective\_agreement} & number \\
        & & \texttt{adjective\_agreement\_strategy} & suffix \\
        & \texttt{comparative} & \texttt{comparative} & comparative, superlative, equative \\
        & & \texttt{comparative\_marking\_strategy} & prepositional word \\
        & \texttt{tense\_aspect} & \texttt{tense\_aspect} & present, past, future, imperfect \\
        & & \texttt{tense\_aspect\_marking\_strategy} & suffix \\
        & \texttt{mood} & \texttt{mood} & indicative, subjunctive, imperative, conditional \\
        & & \texttt{mood\_marking\_strategy} & suffix \\
        & \texttt{voice} & \texttt{voice} & active, passive \\
        & & \texttt{voice\_marking\_strategy} & suffix \\
        & \texttt{person} & \texttt{person\_agreement} & first, second, third \\
        & & \texttt{person\_marking\_strategy} & suffix \\
        & & \texttt{verbal\_number\_agreement} & singular, plural \\
        & & \texttt{verbal\_number\_marking\_strategy} & suffix \\
        & \texttt{inclusive\_exclusive} & & True \\
        & \texttt{nominal\_number} & \texttt{nominal\_number} & singular, plural \\
        & & \texttt{nominal\_number\_marking\_strategy} & suffix \\
        & \texttt{relativization} & \texttt{relativization\_order} & head-initial \\
        & & \texttt{relativization\_marking} & head-marking \\
        & & \texttt{relativizer\_position} & postpositional \\
        & & \texttt{relativizer\_morpheme} & word \\
        & \texttt{negation} & & postpositional word \\
        & \texttt{infinitive} & \texttt{infinitive} & infinitive \\
        & & \texttt{infinitive\_position} & suffix \\
        \bottomrule
    \end{tabular}
    \caption{The configuration of the Hixkaryana-like feature set.
    The configuration is largely based on \citet{Derbyshire:85}.
    }
    \label{tab:hixkaryana-spec}
\end{table*}

\begin{table*}[t]
    \centering
    \scriptsize
    \setlength{\tabcolsep}{5pt}
    \begin{tabular}{@{}lllp{18em}@{}} \toprule
        Submodule & Features & Subfeatures & Value \\
        \midrule
        Syntax & \texttt{main\_word\_order} & & OSV \\
        & \texttt{oblique\_word\_order} & & XOV \\
        & \texttt{adj\_noun\_word\_order} & & NA \\
        & \texttt{posspron\_pron\_word\_order} & & PossN \\
        & \texttt{num\_noun\_word\_order} & & NNum \\
        & \texttt{adposition\_noun\_word\_order} & & NP \\
        \midrule
        Morphology & \texttt{case} & \texttt{case\_marking} & ergative, absolutive, genitive, instrumental \\ \\
        & & \texttt{case\_marking\_strategy} & postpositional word \\
        & & \texttt{oblique\_case\_marking} & None \\
        & \texttt{definiteness} & & None \\
        & \texttt{adjective\_agreement} & & None \\
        & \texttt{comparative} & \texttt{comparative} & comparative, superlative \\
        & & \texttt{comparative\_marking\_strategy} & postpositional word \\
        & \texttt{tense\_aspect} & & None \\
        & \texttt{mood} & & None \\
        & \texttt{voice} & & None \\
        & \texttt{person} & \texttt{person\_agreement} & first, second, third \\
        & & \texttt{person\_marking\_strategy} & prepositional word \\
        & & \texttt{verbal\_number\_agreement} & singular, plural \\
        & & \texttt{verbal\_number\_marking\_strategy} & prepositional word \\
        & \texttt{inclusive\_exclusive} & & False \\
        & \texttt{nominal\_number} & & None \\
        & \texttt{relativization} & \texttt{relativization\_order} & head-initial \\
        & & \texttt{relativization\_marking} & dependent-marking \\
        & & \texttt{relativizer\_position} & postpositional \\
        & & \texttt{relativizer\_morpheme} & affix \\
        & \texttt{negation} & & postpositional word \\
        & \texttt{infinitive} & & None \\
        \bottomrule
    \end{tabular}
    \caption{The configuration of the Mizo-like feature set.
    The configuration is largely based on \citet{Chhangte:89}.}
    \label{tab:mizo-spec}
\end{table*}

\begin{table*}[t]
    \centering
    \scriptsize
    \setlength{\tabcolsep}{5pt}
    \begin{tabular}{@{}lllp{18em}@{}} \toprule
        Submodule & Features & Subfeatures & Value \\
        \midrule
        Syntax & \texttt{main\_word\_order} & & SOV \\
        & \texttt{oblique\_word\_order} & & XOV \\
        & \texttt{adj\_noun\_word\_order} & & AN \\
        & \texttt{posspron\_pron\_word\_order} & & PossN \\
        & \texttt{num\_noun\_word\_order} & & NumN \\
        & \texttt{adposition\_noun\_word\_order} & & NP \\
        \midrule
        Morphology & \texttt{case} & \texttt{case\_marking} & nominative, accusative, dative, genitive, ablative, locative, instrumental \\
        & & \texttt{case\_marking\_strategy} & suffix \\
        & & \texttt{oblique\_case\_marking} & genitive \\
        & \texttt{definiteness} & & None \\
        & \texttt{adjective\_agreement} & & None \\
        & \texttt{comparative} & \texttt{comparative} & comparative, superlative \\
        & & \texttt{comparative\_marking\_strategy} & prepositional word \\
        & \texttt{tense\_aspect} & \texttt{tense\_aspect} & present, past, future \\
        & & \texttt{tense\_aspect\_marking\_strategy} & suffix \\
        & \texttt{mood} & \texttt{mood} & indicative, imperative, conditional \\
        & & \texttt{mood\_marking\_strategy} & suffix \\
        & \texttt{voice} & \texttt{voice} & active, passive \\
        & & \texttt{voice\_marking\_strategy} & suffix \\
        & \texttt{person} & \texttt{person\_agreement} & first, second, third \\
        & & \texttt{person\_marking\_strategy} & suffix \\
        & & \texttt{verbal\_number\_agreement} & singular, plural \\
        & & \texttt{verbal\_number\_marking\_strategy} & suffix \\
        & \texttt{inclusive\_exclusive} & & False \\
        & \texttt{nominal\_number} & \texttt{nominal\_number} & singular, plural \\
        & & \texttt{nominal\_number\_marking\_strategy} & suffix \\
        & \texttt{relativization} & \texttt{relativization\_order} & head-final \\
        & & \texttt{relativization\_marking} & dependent-marking \\
        & & \texttt{relativizer\_position} & postpositional \\
        & & \texttt{relativizer\_morpheme} & affix \\
        & \texttt{negation} & & suffix \\
        & \texttt{infinitive} & \texttt{infinitive} & infinitive \\
        & & \texttt{infinitive\_position} & suffix \\
        \bottomrule
    \end{tabular}
    \caption{The configuration of the Turkish-like feature set.}
    \label{tab:turkish-spec}
\end{table*}

\begin{table*}[t]
    \centering
    \scriptsize
    \setlength{\tabcolsep}{5pt}
    \begin{tabular}{@{}lllp{18em}@{}} \toprule
        Submodule & Features & Subfeatures & Value \\
        \midrule
        Syntax & \texttt{main\_word\_order} & & SVO \\
        & \texttt{oblique\_word\_order} & & VOX \\
        & \texttt{adj\_noun\_word\_order} & & NA \\
        & \texttt{posspron\_pron\_word\_order} & & NPoss \\
        & \texttt{num\_noun\_word\_order} & & NumN \\
        & \texttt{adposition\_noun\_word\_order} & & PN \\
        \midrule
        Morphology & \texttt{case} & & None \\
        & \texttt{definiteness} & & None \\
        & \texttt{adjective\_agreement} & & None \\
        & \texttt{comparative} & \texttt{comparative} & comparative, superlative, equative \\
        & & \texttt{comparative\_marking\_strategy} & postpositional word \\
        & \texttt{tense\_aspect} & & None \\
        & \texttt{mood} & & None \\
        & \texttt{voice} & & None \\
        & \texttt{person} & & None \\
        & \texttt{inclusive\_exclusive} & & False \\
        & \texttt{nominal\_number} & & None \\
        & \texttt{relativization} & \texttt{relativization\_order} & head-initial \\
        & & \texttt{relativization\_marking} & head-marking \\
        & & \texttt{relativizer\_position} & postpositional \\
        & & \texttt{relativizer\_morpheme} & word \\
        & \texttt{negation} & & prepositional word \\
        & \texttt{infinitive} & \texttt{infinitive} & None \\
        \bottomrule
    \end{tabular}
    \caption{The configuration of the Vietnamese-like feature set.}
    \label{tab:vietnamese-spec}
\end{table*}

\begin{table*}[t]
    \centering
    \scriptsize
    \setlength{\tabcolsep}{5pt}
    \begin{tabular}{@{}lllp{18em}@{}} \toprule
        Submodule & Features & Subfeatures & Value \\
        \midrule
        Syntax & \texttt{main\_word\_order} & & VSO \\
        & \texttt{oblique\_word\_order} & & VOX \\
        & \texttt{adj\_noun\_word\_order} & & NA \\
        & \texttt{posspron\_pron\_word\_order} & & NPoss \\
        & \texttt{num\_noun\_word\_order} & & NumN \\
        & \texttt{adposition\_noun\_word\_order} & & PN \\
        \midrule
        Morphology & \texttt{case} & & None \\
        & \texttt{definiteness} & \texttt{definiteness} & definite \\
        & & \texttt{definiteness\_marking\_strategy} & prepositional word \\
        & & \texttt{definiteness\_agreement} & None \\
        & \texttt{adjective\_agreement} & & None \\
        & \texttt{comparative} & \texttt{comparative} & comparative, superlative \\
        & & \texttt{comparative\_marking\_strategy} & suffix \\
        & \texttt{tense\_aspect} & \texttt{tense\_aspect} & present, past, future \\
        & & \texttt{tense\_aspect\_marking\_strategy} & suffix \\
        & \texttt{mood} & \texttt{mood} & indicative, subjunctive, imperative, conditional \\
        & & \texttt{mood\_marking\_strategy} & suffix \\
        & \texttt{voice} & & None \\
        & \texttt{person} & \texttt{person\_agreement} & first, second, third \\
        & & \texttt{person\_marking\_strategy} & suffix \\
        & & \texttt{verbal\_number\_agreement} & singular, plural \\
        & & \texttt{verbal\_number\_marking\_strategy} & suffix \\
        & \texttt{inclusive\_exclusive} & & False \\
        & \texttt{nominal\_number} & \texttt{nominal\_number} & singular, plural \\
        & & \texttt{nominal\_number\_marking\_strategy} & suffix \\
        & \texttt{relativization} & \texttt{relativization\_order} & head-initial \\
        & & \texttt{relativization\_marking} & head-marking \\
        & & \texttt{relativizer\_position} & postpositional \\
        & & \texttt{relativizer\_morpheme} & word \\
        & \texttt{negation} & & prepositional word \\
        & \texttt{infinitive} & \texttt{infinitive} & None \\
        \bottomrule
    \end{tabular}
    \caption{The configuration of the Welsh-like feature set.}
    \label{tab:welsh-spec}
\end{table*}

\begin{table*}[t]
    \centering
    \scriptsize
    \setlength{\tabcolsep}{5pt}
    \begin{tabular}{@{}lllp{18em}@{}} \toprule
        Submodule & Features & Subfeatures & Value \\
        \midrule
        Syntax & \texttt{main\_word\_order} & & OSV \\
        & \texttt{oblique\_word\_order} & & OXV \\
        & \texttt{adj\_noun\_word\_order} & & NA \\
        & \texttt{posspron\_pron\_word\_order} & & NPoss \\
        & \texttt{num\_noun\_word\_order} & & NNum \\
        & \texttt{adposition\_noun\_word\_order} & & NP \\
        \midrule
        Morphology & \texttt{case} & \texttt{case\_marking} & ergative, absolutive, genitive, dative, locative, instrumental \\
        & & \texttt{case\_marking\_strategy} & prefix \\
        & & \texttt{oblique\_case\_marking} & instrumental \\
        & \texttt{definiteness} & \texttt{definiteness} & definite, indefinite \\
        & & \texttt{definiteness\_marking\_strategy} & suffix \\
        & \texttt{adjective\_agreement} & \texttt{adjective\_agreement} & number, case, definiteness \\
        & & \texttt{adjective\_agreement\_strategy} & prefix \\
        & \texttt{comparative} & \texttt{comparative} & comparative, superlative, equative \\
        & & \texttt{comparative\_marking\_strategy} & prefix \\
        & \texttt{tense\_aspect} & \texttt{tense\_aspect} & present, future, recent past, remote past \\
        & & \texttt{tense\_aspect\_marking\_strategy} & prefix \\
        & \texttt{mood} & \texttt{mood} & indicative, subjunctive, imperative, conditional \\
        & & \texttt{mood\_marking\_strategy} & prefix \\
        & \texttt{voice} & \texttt{voice} & active, passive \\
        & & \texttt{voice\_marking\_strategy} & prefix \\
        & \texttt{person} & \texttt{person\_agreement} & first, second, third \\
        & & \texttt{person\_marking\_strategy} & suffix \\
        & & \texttt{verbal\_number\_agreement} & singular, plural, dual \\
        & & \texttt{verbal\_number\_marking\_strategy} & prefix \\
        & \texttt{inclusive\_exclusive} & & True \\
        & \texttt{nominal\_number} & \texttt{nominal\_number} & singular, plural, dual \\
        & & \texttt{nominal\_number\_marking\_strategy} & prefix \\
        & \texttt{relativization} & \texttt{relativization\_order} & head-final \\
        & & \texttt{relativization\_marking} & head-marking \\
        & & \texttt{relativizer\_position} & postpositional \\
        & & \texttt{relativizer\_morpheme} & affix \\
        & \texttt{negation} & & suffix \\
        & \texttt{infinitive} & \texttt{infinitive} & infinitive \\
        & & \texttt{infinitive\_position} & prefix \\
        \bottomrule
    \end{tabular}
    \caption{The configuration of the ``hard'' feature set.}
    \label{tab:hard-language-spec}
\end{table*}

\clearpage

\section{Full results of the morphosyntax experiments}\label{sec:morphosyntax-full-results}

Full details of the morphosyntax experiments are presented in Table~\ref{tab:full-results}.

\begin{table*}[t]
\centering
\setlength{\tabcolsep}{3pt}
\tiny
\begin{minipage}[t]{0.49\linewidth}
\begin{tabular}{@{}lllrrrrr@{}} \toprule
    Model & Language & ICL & TER & SER & MFER & MSER & LemF1 \\ \midrule
    claude-3-5-sonnet     & arabic & No & 75.76 & 0.00 & 34.18 & 17.09 & 93.12 \\
    & fijian & No & 56.25 & 30.82 & 33.51 & 32.16 & 63.07 \\
    & french & No & 98.63 & 15.05 & 36.22 & 25.64 & 88.99 \\
    & hixkaryana & No & 61.86 & 15.73 & 42.60 & 29.17 & 84.23 \\
    & mizo & No & 113.13 & 92.15 & 61.71 & 76.93 & 52.58 \\
    & turkish & No & 67.67 & 33.83 & 20.70 & 27.27 & 91.37 \\
    & welsh & No & 52.17 & 15.31 & 21.90 & 18.60 & 91.24 \\
    & vietnamese & No & 75.25 & 31.25 & 48.57 & 39.91 & 58.10 \\
    & hard & No & 99.26 & 66.18 & 76.16 & 71.17 & 62.87 \\
    \cmidrule{2-8}
    & Average & No & 77.78 & 33.37 & 41.73 & 37.55 & 76.17 \\
    \midrule
    gemini-2.5-flash     & arabic & No & 60.61 & 30.93 & 46.30 & 38.61 & 74.40 \\
    & fijian & No & 22.50 & 0.00 & 28.76 & 14.38 & 83.54 \\
    & french & No & 73.97 & 0.00 & 39.05 & 19.53 & 82.68 \\
    & hixkaryana & No & 92.78 & 31.47 & 57.97 & 44.72 & 72.71 \\
    & mizo & No & 75.42 & 61.43 & 47.36 & 54.40 & 65.91 \\
    & turkish & No & 135.34 & 84.59 & 56.34 & 70.46 & 79.40 \\
    & welsh & No & 52.17 & 0.00 & 32.22 & 16.11 & 90.16 \\
    & vietnamese & No & 45.15 & 46.88 & 42.43 & 44.65 & 62.27 \\
    & hard & No & 115.81 & 132.35 & 78.37 & 105.36 & 46.94 \\
    \cmidrule{2-8}
    & Average & No & 74.86 & 43.07 & 47.64 & 45.36 & 73.11 \\
    \midrule
    gemini-2.5-pro     & arabic & No & 60.61 & 15.46 & 38.82 & 27.14 & 87.89 \\
    & fijian & No & 67.50 & 92.47 & 48.80 & 70.63 & 66.56 \\
    & french & No & 86.30 & 15.05 & 44.39 & 29.72 & 81.99 \\
    & hixkaryana & No & 123.71 & 47.20 & 50.30 & 48.75 & 73.02 \\
    & mizo & No & 75.42 & 46.08 & 47.76 & 46.92 & 77.04 \\
    & turkish & No & 135.34 & 84.59 & 56.30 & 70.45 & 77.65 \\
    & welsh & No & 52.17 & 0.00 & 27.54 & 13.77 & 91.40 \\
    & vietnamese & No & 30.10 & 31.25 & 33.85 & 32.55 & 73.74 \\
    & hard & No & 115.81 & 215.07 & 80.33 & 147.70 & 40.14 \\
    \cmidrule{2-8}
    & Average & No & 83.00 & 60.80 & 47.57 & 54.18 & 74.38 \\
    \midrule
    gpt-4o-mini     & arabic & No & 90.91 & 77.32 & 77.70 & 77.51 & 51.66 \\
    & fijian & No & 56.25 & 61.64 & 56.79 & 59.22 & 48.21 \\
    & french & No & 73.97 & 45.15 & 69.12 & 57.13 & 60.17 \\
    & hixkaryana & No & 108.25 & 94.41 & 62.51 & 78.46 & 55.91 \\
    & mizo & No & 87.99 & 92.15 & 62.29 & 77.22 & 54.06 \\
    & turkish & No & 101.50 & 67.67 & 71.06 & 69.36 & 52.06 \\
    & welsh & No & 78.26 & 61.22 & 64.70 & 62.96 & 51.29 \\
    & vietnamese & No & 75.25 & 62.50 & 54.31 & 58.40 & 51.63 \\
    & hard & No & 115.81 & 82.72 & 81.56 & 82.14 & 50.08 \\
    \cmidrule{2-8}
    & Average & No & 87.58 & 71.64 & 66.67 & 69.16 & 52.79 \\
    \midrule
    gpt-5-mini     & arabic & No & 75.76 & 0.00 & 38.72 & 19.36 & 88.14 \\
    & fijian & No & 67.50 & 92.47 & 61.71 & 77.09 & 46.76 \\
    & french & No & 73.97 & 135.45 & 45.20 & 90.32 & 59.74 \\
    & hixkaryana & No & 77.32 & 62.94 & 34.48 & 48.71 & 81.06 \\
    & mizo & No & 100.56 & 61.43 & 60.64 & 61.04 & 50.46 \\
    & turkish & No & 118.42 & 67.67 & 51.48 & 59.58 & 80.49 \\
    & welsh & No & 39.13 & 0.00 & 21.92 & 10.96 & 89.40 \\
    & vietnamese & No & 75.25 & 78.12 & 48.29 & 63.21 & 57.56 \\
    & hard & No & 99.26 & 66.18 & 67.75 & 66.96 & 60.89 \\
    \cmidrule{2-8}
    & Average & No & 80.80 & 62.70 & 47.80 & 55.25 & 68.28 \\
    \midrule
    gpt-5     & arabic & No & 75.76 & 15.46 & 35.65 & 25.55 & 83.98 \\
    & fijian & No & 67.50 & 77.05 & 60.00 & 68.53 & 45.97 \\
    & french & No & 73.97 & 15.05 & 39.33 & 27.19 & 76.54 \\
    & hixkaryana & No & 30.93 & 15.73 & 35.47 & 25.60 & 80.85 \\
    & mizo & No & 100.56 & 76.79 & 63.14 & 69.97 & 46.99 \\
    & turkish & No & 135.34 & 67.67 & 61.46 & 64.57 & 79.90 \\
    & welsh & No & 78.26 & 45.92 & 48.36 & 47.14 & 74.32 \\
    & vietnamese & No & 45.15 & 46.88 & 46.74 & 46.81 & 58.78 \\
    & hard & No & 99.26 & 49.63 & 26.20 & 37.91 & 8.49 \\
    \cmidrule{2-8}
    & Average & No & 78.53 & 45.58 & 46.26 & 45.92 & 61.76 \\
    \bottomrule
\end{tabular}
\end{minipage}
\begin{minipage}[t]{0.49\linewidth}
\begin{tabular}{@{}lllrrrrr@{}} \toprule
    Model & Language & ICL & TER & SER & MFER & MSER & LemF1 \\ \midrule
    claude-3-5-sonnet     & arabic & Yes & 60.61 & 0.00 & 26.57 & 13.28 & 97.48 \\
    & fijian & Yes & 22.50 & 15.41 & 13.32 & 14.36 & 81.49 \\
    & french & Yes & 73.97 & 15.05 & 19.54 & 17.30 & 93.85 \\
    & hixkaryana & Yes & 77.32 & 47.20 & 44.70 & 45.95 & 87.32 \\
    & mizo & Yes & 87.99 & 76.79 & 31.79 & 54.29 & 80.61 \\
    & turkish & Yes & 50.75 & 16.92 & 10.43 & 13.67 & 96.76 \\
    & welsh & Yes & 26.09 & 0.00 & 8.97 & 4.49 & 96.66 \\
    & vietnamese & Yes & 15.05 & 15.62 & 18.30 & 16.96 & 84.29 \\
    & hard & Yes & 82.72 & 49.63 & 43.97 & 46.80 & 83.74 \\
    \cmidrule{2-8}
    & Average & Yes & 55.22 & 26.29 & 24.18 & 25.23 & 89.13 \\
    \midrule
    gemini-2.5-flash     & arabic & Yes & 60.61 & 15.46 & 26.45 & 20.96 & 84.76 \\
    & fijian & Yes & 11.25 & 0.00 & 17.67 & 8.84 & 87.52 \\
    & french & Yes & 24.66 & 15.05 & 20.09 & 17.57 & 93.45 \\
    & hixkaryana & Yes & 30.93 & 0.00 & 34.02 & 17.01 & 85.42 \\
    & mizo & Yes & 37.71 & 0.00 & 17.18 & 8.59 & 90.10 \\
    & turkish & Yes & 101.50 & 50.75 & 31.56 & 41.15 & 93.21 \\
    & welsh & Yes & 39.13 & 0.00 & 22.42 & 11.21 & 96.99 \\
    & vietnamese & Yes & 30.10 & 31.25 & 12.67 & 21.96 & 88.14 \\
    & hard & Yes & 99.26 & 49.63 & 35.20 & 42.42 & 90.32 \\
    \cmidrule{2-8}
    & Average & Yes & 48.35 & 18.02 & 24.14 & 21.08 & 89.99 \\
    \midrule
    gemini-2.5-pro     & arabic & Yes & 30.30 & 0.00 & 13.80 & 6.90 & 95.64 \\
    & fijian & Yes & 11.25 & 15.41 & 24.82 & 20.11 & 66.63 \\
    & french & Yes & 24.66 & 0.00 & 8.59 & 4.29 & 96.86 \\
    & hixkaryana & Yes & 92.78 & 31.47 & 29.94 & 30.71 & 82.77 \\
    & mizo & Yes & 12.57 & 0.00 & 9.33 & 4.67 & 93.26 \\
    & turkish & Yes & 84.59 & 33.83 & 18.36 & 26.10 & 92.68 \\
    & welsh & Yes & 13.04 & 0.00 & 4.14 & 2.07 & 97.62 \\
    & vietnamese & Yes & 0.00 & 0.00 & 12.74 & 6.37 & 90.55 \\
    & hard & Yes & 99.26 & 33.09 & 28.65 & 30.87 & 87.64 \\
    \cmidrule{2-8}
    & Average & Yes & 40.94 & 12.64 & 16.71 & 14.68 & 89.30 \\
    \midrule
    gpt-4o-mini     & arabic & Yes & 90.91 & 77.32 & 79.23 & 78.28 & 50.12 \\
    & fijian & Yes & 67.50 & 46.23 & 57.76 & 52.00 & 45.10 \\
    & french & Yes & 73.97 & 45.15 & 59.53 & 52.34 & 72.37 \\
    & hixkaryana & Yes & 108.25 & 94.41 & 63.22 & 78.81 & 54.59 \\
    & mizo & Yes & 87.99 & 92.15 & 62.29 & 77.22 & 54.06 \\
    & turkish & Yes & 101.50 & 67.67 & 70.23 & 68.95 & 53.08 \\
    & welsh & Yes & 78.26 & 61.22 & 71.57 & 66.40 & 47.68 \\
    & vietnamese & Yes & 75.25 & 62.50 & 52.37 & 57.43 & 53.57 \\
    & hard & Yes & 115.81 & 82.72 & 81.03 & 81.88 & 50.83 \\
    \cmidrule{2-8}
    & Average & Yes & 88.83 & 69.93 & 66.36 & 68.14 & 53.49 \\
    \midrule
    gpt-5-mini     & arabic & Yes & 45.45 & 0.00 & 25.46 & 12.73 & 91.07 \\
    & fijian & Yes & 101.25 & 138.70 & 69.42 & 104.06 & 37.02 \\
    & french & Yes & 61.64 & 90.30 & 38.50 & 64.40 & 66.20 \\
    & hixkaryana & Yes & 108.25 & 47.20 & 45.44 & 46.32 & 79.01 \\
    & mizo & Yes & 125.70 & 92.15 & 64.11 & 78.13 & 46.30 \\
    & turkish & Yes & 118.42 & 84.59 & 51.38 & 67.98 & 80.18 \\
    & welsh & Yes & 26.09 & 0.00 & 17.76 & 8.88 & 92.07 \\
    & vietnamese & Yes & 60.20 & 62.50 & 47.23 & 54.86 & 57.51 \\
    & hard & Yes & 115.81 & 49.63 & 54.81 & 52.22 & 72.80 \\
    \cmidrule{2-8}
    & Average & Yes & 84.76 & 62.79 & 46.01 & 54.40 & 69.13 \\
    \midrule
    gpt-5     & arabic & Yes & 60.61 & 15.46 & 26.46 & 20.96 & 91.59 \\
    & fijian & Yes & 33.75 & 15.41 & 38.24 & 26.83 & 66.67 \\
    & french & Yes & 24.66 & 0.00 & 10.38 & 5.19 & 96.10 \\
    & hixkaryana & Yes & 46.39 & 31.47 & 33.82 & 32.64 & 82.19 \\
    & mizo & Yes & 37.71 & 15.36 & 25.53 & 20.45 & 79.47 \\
    & turkish & Yes & 118.42 & 84.59 & 51.27 & 67.93 & 79.38 \\
    & welsh & Yes & 52.17 & 45.92 & 38.47 & 42.19 & 80.22 \\
    & vietnamese & Yes & 45.15 & 46.88 & 28.77 & 37.82 & 78.17 \\
    & hard & Yes & 99.26 & 66.18 & 52.20 & 59.19 & 65.35 \\
    \cmidrule{2-8}
    & Average & Yes & 57.57 & 35.70 & 33.90 & 34.80 & 79.90 \\
    \bottomrule
\end{tabular}
\end{minipage}
\caption{Full results of the Morphosyntax module.
The left table reports the performance without the ``review'' stage for few-shot in-context learning, and the right table reports the performance with the ``review'' stage.}
\label{tab:full-results}
\end{table*}

\end{document}